\documentclass[letterpaper]{article} 
\usepackage{aaai24} 
\usepackage{times}  
\usepackage{helvet}  
\usepackage{courier}  
\usepackage[hyphens]{url}  
\usepackage{graphicx} 
\usepackage{natbib}  
\usepackage{caption} 
\usepackage{algorithm}
\usepackage{algorithmic}
\usepackage{amsmath}
\usepackage{amsfonts} 
\usepackage{newfloat}
\usepackage{listings}
\usepackage{adjustbox}

\usepackage{xspace}
\usepackage{multirow}
\usepackage[table,xcdraw]{xcolor}

\DeclareCaptionStyle{ruled}{labelfont=normalfont,labelsep=colon,strut=off} 
\floatstyle{ruled}
\newfloat{listing}{tb}{lst}{}
\floatname{listing}{Listing}
\makeatletter
\DeclareRobustCommand\onedot{\futurelet\@let@token\@onedot}
\def\@onedot{\ifx\@let@token.\else.\null\fi\xspace}

\def\etc{\emph{etc}\onedot} 
 
\def\etal{\emph{et al}\onedot}
\makeatother

\title{MedPrompt: Cross-Modal Prompting for Multi-Task Medical Image Translation}
\author {
    Xuhang Chen\textsuperscript{\rm 12},
    Chi-Man Pun\textsuperscript{\rm 2}\footnotemark[1],
    Shuqiang Wang\textsuperscript{\rm 1}\thanks{Corresponding Author}
}
\affiliations {
    \textsuperscript{\rm 1}Shenzhen Institutes of Advanced Technology, Chinese Academy of Sciences\\
    \textsuperscript{\rm 2}University of Macau\\
    
}

\begin{document}
\nocopyright
\maketitle

\begin{abstract}
Cross-modal medical image translation is an essential task for synthesizing missing modality data for clinical diagnosis. However, current learning-based techniques have limitations in capturing cross-modal and global features, restricting their suitability to specific pairs of modalities. This lack of versatility undermines their practical usefulness, particularly considering that the missing modality may vary for different cases.
In this study, we present MedPrompt, a multi-task framework that efficiently translates different modalities. Specifically, we propose the Self-adaptive Prompt Block, which dynamically guides the translation network towards distinct modalities. Within this framework, we introduce the Prompt Extraction Block and the Prompt Fusion Block to efficiently encode the cross-modal prompt. To enhance the extraction of global features across diverse modalities, we incorporate the Transformer model. Extensive experimental results involving five datasets and four pairs of modalities demonstrate that our proposed model achieves state-of-the-art visual quality and exhibits excellent generalization capability.

\end{abstract}

\section{Introduction}

Multi-modal medical images play a crucial role in precision medicine and public health studies~\cite{brody2013medical} since each modality provides unique anatomical or functional information about the human body, which refer to medical imaging data acquired from multiple distinct imaging modalities, such as Computed Tomography (CT), Magnetic Resonance Imaging (MRI), Positron Emission Tomography (PET) scan, \etc. Each imaging modality provides different information and perspectives. By combining the imaging data from multiple modalities, a more comprehensive, accurate, and detailed representation of the medical condition or anatomy can be obtained. However, the widespread implementation of multi-modal imaging faces various challenges, such as patient non-compliance and lengthy scan durations. Consequently, cross-modal medical image translation has gained popularity due to its low-cost nature and ability to identify disease areas, facilitate precise and early diagnosis, and serve various purposes like super-resolution \etc~\cite{you2022fine,hu20233,wang2012bayesian,lei2022longitudinal}.



\begin{figure}[t]
    \begin{minipage}[b]{1.0\linewidth}
        \begin{minipage}[b]{.32\linewidth}
            \centering
            \centerline{\includegraphics[width=\linewidth]{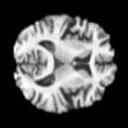}}
            \centerline{(a) Input}\medskip
        \end{minipage}
        \hfill
        \begin{minipage}[b]{.32\linewidth}
            \centering
            \centerline{\includegraphics[width=\linewidth]{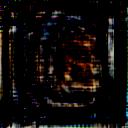}}
            \centerline{(b) CycleGAN}\medskip
        \end{minipage}
        \hfill
        \begin{minipage}[b]{0.32\linewidth}
            \centering
            \centerline{\includegraphics[width=\linewidth]{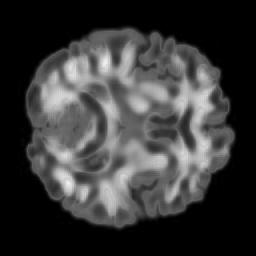}}
            \centerline{(c) medSynth}\medskip
        \end{minipage}
    \end{minipage}
    \begin{minipage}[b]{1.0\linewidth}
        \begin{minipage}[b]{.32\linewidth}
            \centering
            \centerline{\includegraphics[width=\linewidth]{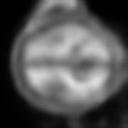}}
            \centerline{(d) pGAN}\medskip
        \end{minipage}
        \hfill
        \begin{minipage}[b]{.32\linewidth}
            \centering
            \centerline{\includegraphics[width=\linewidth]{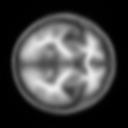}}
            \centerline{(e) MedPrompt}\medskip
        \end{minipage}
        \hfill
        \begin{minipage}[b]{0.32\linewidth}
            \centering
            \centerline{\includegraphics[width=\linewidth]{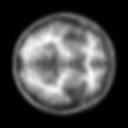}}
            \centerline{(f) Target}\medskip
        \end{minipage}
    \end{minipage}
    \caption{
    The visual results consist of (a) the input MRI, (b) the CycleGAN~\cite{zhu2017unpaired} output, (c) the medSynth~\cite{nie2017medical} output, (d) the pGAN~\cite{dar2019image} output, (e) the MedPrompt output, and (f) the target PET. Our result demonstrates superior visual quality and similarity when compared to the other methods.
    } 
    \label{fig:teaser}
\end{figure}

However, translating cross-modal medical images poses a challenging inverse problem due to their high dimensionality and nonlinear variations in tissue contrast across different modalities~\cite{huang2017cross}. General image translation models designed for natural images often struggle to capture the specific features and characteristics of medical modalities, as seen in Figure~\ref{fig:teaser} (b). 

There exist specialized deep learning models for medical image translation~\cite{nie2017medical,dar2019image}. Despite their effectiveness, most medical image translation models heavily depend on convolutional frameworks that use compact filters for local image feature extraction. These frameworks often fail to capture contextual features that represent long-range spatial dependencies, as they primarily focus on small pixel neighborhoods, as shown in Figure~\ref{fig:teaser} (c) and (d). Although ResViT~\cite{dalmaz2022resvit} attempted to address these limitations by incorporating the Vision Transformer~\cite{dosovitskiy2020vit}, it still exhibits limitations in terms of generalization capability and performance across different modalities.

To address the aforementioned challenges, we propose MedPrompt, a cross-modal Transformer based on prompting for multi-task medical image translation. MedPrompt leverages the Transformer architecture to extract global features from diverse modalities, benefiting from its wide receptive field. We employ the technique of prompting~\cite{jia2022visual}, which utilizes adjustable parameters to encode vital differentiating information specific to each medical image modality. This approach empowers the model to capture a wide range of cross-modal feature pairs and improves its adaptability.

The main contributions of our work are as follows:
\begin{enumerate}
    \item We propose a simple but novel Self-adaptive Prompt Block, in which we introduce a Prompt Extraction Block and a Prompt Fusion Block to effectively encode and aggregate cross-modal prompt.
    
    \item Due to the cross-modal features provided by the Self-adaptive Prompt Block and the global receptive field offered by the Transformer, our model demonstrates promising performance in multi-task medical image translation. These features enable our model to effectively capture and utilize information from different modalities, leading to improved translation results.
    
    \item Extensive experiments demonstrate the effectiveness of the proposed model through both quantitative and qualitative results.
\end{enumerate}

\section{Related Work}

\subsection{Image-to-Image Translation}
Image-to-Image Translation is a significant task that aims to learn a mapping between an input image and an output image. 
CycleGAN~\cite{zhu2017unpaired} establishes a cycle-consistency invariant, allowing it to learn the mapping between two domains without requiring a large number of aligned image pairs. 
Pix2Pix~\cite{isola2017image} is a GAN-based model that maps input pixel space to target pixel space at the pixel level. 
UNIT~\cite{liu2017unsupervised} regards the image translation problem as learning the joint probability density, with each data space sharing a latent space. 
MUNIT~\cite{huang2018multimodal} highlights the presence of a separate space referred to as the style space, which captures the variations and distinctions among these domains.
FUNIT~\cite{liu2019few} introduces a few-shot learning approach that leverages the decomposition of the content space and the style space to capture style information from a small set of reference images, enabling image translation.	
U-GAT-IT~\cite{kim2019u} is a novel unsupervised image-to-image translation method that combines a new attention module and a learnable normalization function in an end-to-end manner.
CUT~\cite{park2020contrastive} is an image translation method based on contrastive learning. It utilizes the effectiveness of contrastive learning techniques and discovers that extracting negative image patches from a single image yields better results compared to extracting from other images in the dataset.
LPTN~\cite{liang2021high} is a lightweight image translation method for high-resolution images based on Laplacian Pyramid.

\subsection{Cross-modal Medical Image Translation}
In recent years, deep learning models have enabled rapid developments in cross-modal medical Image translation. 
The medSynth~\cite{nie2017medical} initiates the process of medical image synthesis using Deep Convolutional Adversarial Networks.
RIED-Net~\cite{gao2019deep} introduces a method that aims to learn the nonlinear mapping between MRI inputs and targeted PET images.
Dar \etal introduce pGAN~\cite{dar2019image} as a method for enhancing the accuracy of synthesized multi-contrast MRI images.
BMGAN~\cite{hu2021bidirectional} focuses on the bidirectional mapping between Brain MRI and PET modalities using generative adversarial networks.
ResViT~\cite{dalmaz2022resvit} introduces Residual Vision Transformers for cross-modal medical image synthesis.


\begin{figure*}[ht]
    \begin{minipage}[b]{1.0\linewidth}
        \begin{minipage}[b]{1.0\linewidth}
            \centering
            \centerline{\includegraphics[width=\linewidth]{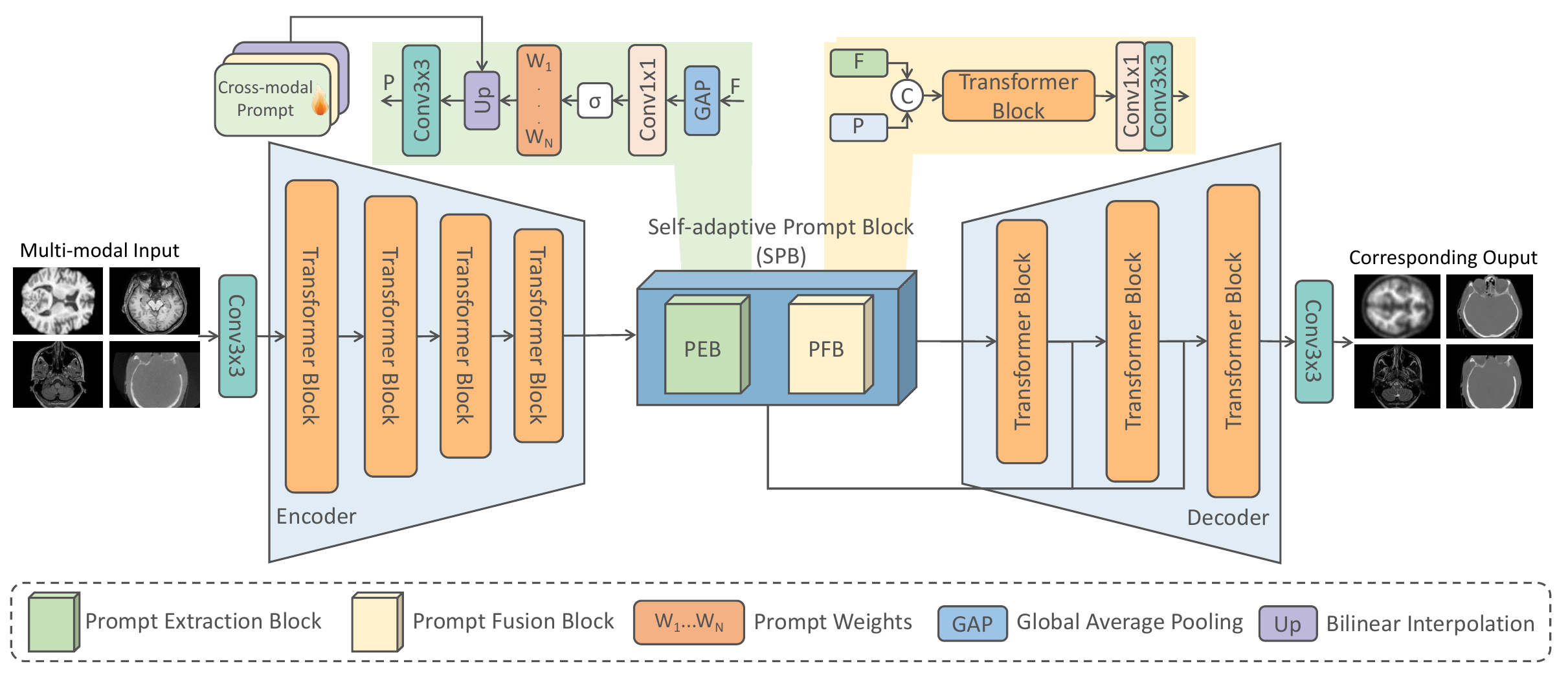}}
        \end{minipage}
    \end{minipage}
    \caption{
    The overall pipeline of our MedPrompt. We employ a typical encoder-decoder framework. Given a cross-modal dataset, we input all the distinct modalities at the beginning of training. Self-adaptive Prompt Block (SPB) is introduced after the 4-level encoder. We propose Prompt Extraction Block (PEB) and Prompt Fusion Block (PFB) to encode and aggregate prompt information from multiple modalities. From the last layer of the encoder, each SPB connects the preceding and succeeding transformer blocks. In this way, each prompt information is propagated between decoders, eventually generating pleasing results.
    } 
    \label{fig:model}
\end{figure*}

\section{Methodology}

\subsection{Overview}
MedPrompt is a classical encoder-decoder architecture model. Given a set of multi-modal input images $D_{L}=\left\{\left(x_{i}^{l}, y_{i}^{l}\right) \mid x_{i}^{l} \in \mathcal{I}_{s}^{IN}, y_{i}^{l} \in \mathcal{I}_{s}^{GT}\right\}_{i=1}^{N}$, where $x_{i}^{l}$ and $y_{i}^{l}$ are the input image set $\mathcal{I}_{s}^{IN}$ and groudtruth set $\mathcal{I}_{s}^{GT}$. MedPrompt first extracts low-level features by a $3 \times 3$ convolution and outputs features $F_{0}$, which are then fed into a 4-level encoder. Each encoder level consists of gradually increasing Transformer blocks. 

The key contribution of our work lies in the proposed simple prompt-based approach for multi-task medical image translation. Therefore, in our proposed MedPromt framework, we utilize an existing Transformer encoder block as the basic architecture from~\cite{zamir2022restormer}, rather than developing a new one specifically for this task. Each Transformer block in the proposed framework contains a Multi-Dconv Head Transposed Attention (MDTA) and a Gated-Dconv Feed-Forward Network (GDFN). To better learn the cross-modal prompt from distinct modalities, we propose a Self-adaptive Prompt Block (SPB) composed of Prompt Extraction Block (PEB) and Prompt Fusion Block (PFB), where PEB encode the cross-modal prompt and PFB aggregate the cross-modal prompt. From the last layer transformer block of the encoder, SPBs are inserted between the preceding and succeeding transformer blocks, allowing the cross-modal prompt to propagate between each decoder. 

\subsection{Self-adaptive Prompt Block}
The prompting technique first came from NLP~\cite{houlsby2019parameter,sanh2021multitask,brown2020language,li2021prefix}. In recent years, visual prompting started to demonstrate its efficiency~\cite{jia2022visual,khattak2023maple,sohn2023visual}. There have been efforts investigating prompt-engineering approaches for fine-tuning pre-trained models in a data-efficient manner, with the aim of adapting large frozen models pre-trained on a source task $A$ for the optimization of a distinct target task $B$'s objective. The promise that underpins prompt-engineering approaches stems from their potential in compactly seeding task-specific contextual cues within the prompts, which helps to optimally guide the pre-trained model's optimization towards the target task objective. 

Building upon insights from prior work, we propose a novel multi-task framework for medical image translation, where the most important component is the Self-adaptive Prompt Block (SPB). The SPB parameterizes the prompts as learnable embeddings that can efficiently extract and interact with the input features, with the aim of augmenting them with task-specific information regarding the modality type. Consider input feature as $F \in \mathbb{R}^{C \times H \times W}$, a set of N cross-modal prompt as $P\in \mathbb{R}^{N \times C \times H \times W}$ and the output feature $\hat{F}$ the SPB can be wrote as Equation~\ref{eq:spb}:

\begin{equation}
    \operatorname{\hat{F} = PFB(PEB(P, F), F)}
    \label{eq:spb}
\end{equation}

\subsubsection{Prompt Extraction Block}
Cross-modal prompt information can interact with input features to generate different modality information. Instead of static prompt components, we propose learnable prompt embeddings that can interact dynamically with the input features. Rather than simply calibrating the features using the learned prompts, our proposed module Prompt Extraction Block (PEB) predicts prompt weights $W_1 ... W_N$ conditioned on the input content and applies them to dynamically gate the prompt components. This input-conditioned gating aims to generate prompts that are more tailored to the specific degradation characteristics in each input. Furthermore, PEB constructs a shared latent space to encourage correlated knowledge sharing across the learnable prompt embeddings.

The PEB is designed to extract input-conditioned prompt weights from the input features. First, we apply global average pooling across the spatial dimensions to obtain a channel-wise feature vector. We then employ a channel downscaling convolution layer followed by a softmax operation to produce the prompt weights. These weights are used to gate the prompt components by adjusting their activations. Finally, a 3x3 convolution layer is applied. In summary, given cross-modal prompt $P_c$, prompt weights $w_i$, input feature $F$ and the final prompt $P$,the PEB process can be described as Equation~\ref{eq:peb}:

\begin{equation}
    \begin{aligned}
        &\mathbf{P}=\operatorname{Conv}_{3 \times 3}\left(\sum_{c=1}^{N} w_{i} \mathbf{P}_{c}\right),\\
        &w_{i}=\operatorname{Softmax}\left(\operatorname{Conv}_{1 \times 1}\left(\operatorname{GAP}\left(\mathbf{F}\right)\right)\right)
    \end{aligned}
    \label{eq:peb}
\end{equation}
where $\operatorname{Conv}_{3 \times 3}$ is the $3 \times 3$ convolution, $\operatorname{GAP}$ means Global Average Pooling.
\subsubsection{Prompt Fusion Block}
As mentioned above, PEB extracts the prompt, thus PFB aims to combine the prompt with the input features. The key objective of our Prompt Fusion Block (PFB) is to allow for information exchange between the input features and the prompts, in order to guide the translation process. In PFB, we first concatenate the input features and generated prompts along the channel dimension, combining their representations. We then apply a Transformer encoder block to this concatenated input. The Transformer helps exploit the degradation information encoded in the prompts to transform the input features in a guided manner. The whole PFB process can be described as Equation~\ref{eq:pfb}:

\begin{equation}
    \hat{\mathbf{F}}=\operatorname{Conv}_{3 \times 3}\left(\operatorname{GDFN}\left(\operatorname{MDTA}\left[\mathbf{F} ; \mathbf{P}\right]\right)\right.
    \label{eq:pfb}
\end{equation}
where MDTA and GDFN are the key components of Restormer~\cite{zamir2022restormer}.

\subsection{Objective Function}
\label{sec:loss}
For model training, we utilize two loss functions: Mean Squared Error Loss $L_{MSE}$ and Structural Similarity Index Loss $L_{SSIM}$. These loss functions play a crucial role in preserving and translating modal details. $L_{MSE}$ measures the average squared difference between the generated image and the target modality, providing a measure of pixel-level fidelity. On the other hand, $L_{SSIM}$ evaluates the structural similarity between the generated image and the modality, capturing perceptual differences beyond mere pixel-level comparison. By incorporating both losses, our model can effectively preserve and translate intricate image details during the training process. The total objective function $L_{total}$ can be represented as:
\begin{equation}
L_{total} = L_{MSE} + \lambda * L_{SSIM},
\label{eq:total}
\end{equation}
where we set the weight $\lambda$ of $L_{SSIM}$ to 0.4 empirically.

\section{Experiments}

\subsection{Experiment Settings}

\subsubsection{Dataset}


In our experiments, we conduct comparisons between MedPrompt and other methods using five datasets and four pairs of modalities. The details of these datasets are as follows:

\begin{enumerate}
    \item ADNI~\cite{zuo2021prior} - This medical imaging dataset focuses on Alzheimer's disease and consists of paired MRI and PET brain images. To ensure consistency, we follow the same preprocessing procedure as BMGAN~\cite{hu2020brain} for image preprocessing.
    \item SynthRAD2023~\cite{synthrad2023} - This medical imaging dataset is structured into two tasks. Task 1 involves MRI to CT image synthesis and includes MRI/CT image pairs. Task 2 focuses on Cone-Beam Computed Tomography (CBCT) to CT image translation and includes CBCT/CT image pairs. The dataset contains two anatomical regions: the brain and the pelvis. We follow the official guidance of SynthRAD2023 for image registration and preprocessing.
    \item IXI - This dataset comprises $\rm T_{1}$-weighted and $\rm T_{2}$-weighted brain MRI images. To preprocess the images, we follow the same procedure as pGAN~\cite{dar2019image}.
    \item BraTS2020~\cite{brats} - This dataset includes $\rm T_{1}$-weighted and $\rm T_{2}$-weighted brain MRI images. We utilize the preprocessing procedure of pGAN~\cite{dar2019image} for this dataset as well.
\end{enumerate}

The details regarding the number of training/testing samples and the resolution of the aforementioned datasets are summarized in Table~\ref{table:dataset}.

\begin{table}[ht]
\adjustbox{width=\columnwidth}{
\begin{tabular}{lccc}
\hline
Dataset            & \# of Training & \# of Testing & Resolution \\ \hline
IXI                & 2275           & 910           & $256 \times 256$        \\
BraTS2020          & 7380           & 2500          & $256 \times 256$        \\
SynthRAD2023 Task1 & 981            & 99            & $256 \times 256$        \\
SynthRAD2023 Task2 & 933            & 147           & $256 \times 256$        \\
ADNI               & 597            & 90            & $128 \times 128$        \\ \hline
\end{tabular}}
\caption{Training/testing samples and resolution of the datasets used in the experiments.}
\label{table:dataset}
\end{table}
\begin{table}[b]
\adjustbox{width=\columnwidth}{
\begin{tabular}{l|cccccc}
\hline
\multirow{3}{*}{Method} & \multicolumn{6}{c}{ADNI}                                                                                                   \\ \cline{2-7} 
                        & \multicolumn{3}{c|}{$\rm MRI\rightarrow PET$}                          & \multicolumn{3}{c}{$\rm PET\rightarrow MRI$}      \\
                        & PSNR$\uparrow$ & SSIM$\uparrow$ & \multicolumn{1}{c|}{MAE$\downarrow$} & PSNR$\uparrow$ & SSIM$\uparrow$ & MAE$\downarrow$ \\ \hline
U-Net                   & \underline{21.27}    & \underline{0.73}     & \multicolumn{1}{c|}{\underline{14.77}}     & 18.09          & \underline{0.66}     & 17.79           \\
CycleGAN                & 8.65           & 0.16           & \multicolumn{1}{c|}{66.62}           & 7.79           & 0.24           & 69.19           \\
Pix2Pix                 & 11.43          & 0.34           & \multicolumn{1}{c|}{46.44}           & 8.98           & 0.38           & 56.51           \\
UNIT                    & 13.21          & 0.38           & \multicolumn{1}{c|}{34.04}           & 10.45          & 0.46           & 47.67           \\
MUNIT                   & 11.57          & 0.35           & \multicolumn{1}{c|}{46.06}           & 11.54          & 0.45           & 39.52           \\
FUNIT                   & 13.73          & 0.30            & \multicolumn{1}{c|}{36.49}           & 11.76          & 0.28           & 40.04           \\
U-GAT-IT                  & 17.26          & 0.39           & \multicolumn{1}{c|}{24.62}           & 13.39          & 0.38           & 34.35           \\
CUT                     & 19.05          & 0.51           & \multicolumn{1}{c|}{20.28}           & 12.32          & 0.33           & 37.28           \\
LPTN                    & 14.88          & 0.30            & \multicolumn{1}{c|}{31.01}           & 12.58          & 0.37           & 34.97           \\
medSynth                & 15.26          & 0.40            & \multicolumn{1}{c|}{26.49}           & 12.51          & 0.13           & 38.95           \\
pGAN                    & 14.78          & 0.35           & \multicolumn{1}{c|}{34.28}           & 15.86          & 0.53           & 24.62           \\
RIED-Net                & 20.72          & 0.68           & \multicolumn{1}{c|}{15.69}           & \underline{18.17}    & \underline{0.66}     & \underline{16.93}     \\
ResViT                  & 20.16          & 0.66           & \multicolumn{1}{c|}{16.57}           & 17.27          & 0.63           & 18.84           \\
\textbf{Ours}                    & \textbf{24.43} & \textbf{0.84}  & \multicolumn{1}{c|}{\textbf{9.73}}   & \textbf{21.00}    & \textbf{0.79}  & \textbf{12.39}  \\ \hline
\end{tabular}}
\caption{Quantitative evaluation on ADNI dataset. The best performance is marked in bold, while the second-best performance is underlined.}
\end{table}

\begin{table*}[ht]
\adjustbox{width=\linewidth}{
\begin{tabular}{l|cccccc|cccccc}
\hline
                         & \multicolumn{6}{c|}{IXI}                                                                                                                                                                                                                                      & \multicolumn{6}{c}{BraTS2020}                                                                                                                                                                                                                                 \\ \cline{2-13} 
                         & \multicolumn{3}{c|}{$\rm T_{1}\rightarrow T_{2}$}                                                                                                & \multicolumn{3}{c|}{$\rm T_{2}\rightarrow T_{1}$}                                                                            & \multicolumn{3}{c|}{$\rm T_{1}\rightarrow T_{2}$}                                                                                                 & \multicolumn{3}{c}{$\rm T_{2}\rightarrow T_{1}$}                                                                            \\ 
\multirow{-3}{*}{Method} & PSNR$\uparrow$                        & SSIM$\uparrow$                      & \multicolumn{1}{c|}{MAE$\downarrow$}                      & PSNR$\uparrow$                        & SSIM$\uparrow$                       & MAE$\downarrow$                      & PSNR$\uparrow$                        & SSIM$\uparrow$                       & \multicolumn{1}{c|}{MAE$\downarrow$}                      & PSNR$\uparrow$                       & SSIM$\uparrow$                       & MAE$\downarrow$                      \\ \hline
U-Net                    & \underline{28.18}                                 & \underline{0.88}                                & \multicolumn{1}{c|}{\underline{4.37}}                                 & 28.25                                 & \underline{0.90}                                  & \underline{4.49}                                 & 24.48                                 & \underline{0.89}                                 & \multicolumn{1}{c|}{7.82}                                 & \underline{25.53}                                & \underline{0.92}                                 & \underline{7.18}                                 \\
CycleGAN                 & 16.01                                 & 0.45                                & \multicolumn{1}{c|}{21.84}                                & 14.43                                 & 0.53                                 & 27.91                                & 14.48                                 & 0.64                                 & \multicolumn{1}{c|}{25.27}                                & 12.84                                & 0.64                                 & 32.21                                \\
Pix2Pix                  & 16.72                                 & 0.53                                & \multicolumn{1}{c|}{19.35}                                & 14.09                                 & 0.52                                 & 28.93                                & 14.66                                 & 0.64                                 & \multicolumn{1}{c|}{24.30}                                 & 12.86                                & 0.63                                 & 31.93                                \\
UNIT                     & 16.80                                  & 0.54                                & \multicolumn{1}{c|}{19.16}                                & 14.16                                 & 0.53                                 & 28.6                                 & 8.07                                  & 0.01                                 & \multicolumn{1}{c|}{70.31}                                & 12.96                                & 0.65                                 & 30.83                                \\
MUNIT                    & 17.27                                 & 0.53                                & \multicolumn{1}{c|}{18.70}                                 & 14.28                                 & 0.54                                 & 28.42                                & 14.93                                 & 0.64                                 & \multicolumn{1}{c|}{24.00}                                   & 15.51                                & 0.65                                 & 21.87                                \\
FUNIT                    & 7.06                                  & 0.07                                & \multicolumn{1}{c|}{109.29}                               & 7.50                                   & 0.11                                 & 99.99                                & 7.19                                  & 0.13                                 & \multicolumn{1}{c|}{104.94}                               & 7.37                                 & 0.17                                 & 99.58                                \\
U-GAT-IT                   & 24.85                                 & 0.79                                & \multicolumn{1}{c|}{6.80}                                  & 26.7                                  & 0.86                                 & 5.55                                 & 24.08                                 & 0.87                                 & \multicolumn{1}{c|}{7.91}                                 & 23.18                                & 0.87                                 & 9.92                                 \\
CUT                      & 18.08                                 & 0.59                                & \multicolumn{1}{c|}{13.87}                                & 19.39                                 & 0.56                                 & 14.14                                & 12.01                                 & 0.12                                 & \multicolumn{1}{c|}{43.12}                                & 22.17                                & 0.80                                  & 10.46                                \\
LPTN                     & 18.93                                 & 0.64                                & \multicolumn{1}{c|}{12.37}                                & 22.27                                 & 0.67                                 & 9.51                                 & 18.93                                 & 0.72                                 & \multicolumn{1}{c|}{12.97}                                & 19.60                                 & 0.75                                 & 13.72                                \\
medSynth                 & 26.72                                 & 0.85                                & \multicolumn{1}{c|}{5.59}                                 & \underline{28.28}                                 & \underline{0.90}                                  & 4.65                                 & 24.08                                 & \underline{0.89}                                 & \multicolumn{1}{c|}{8.78}                                 & 24.73                                & 0.91                                 & 7.95                                 \\
pGAN                     & 25.15                                 & 0.78                                & \multicolumn{1}{c|}{6.80}                                  & 25.58                                 & 0.80                                  & 6.77                                 & 23.34                                 & 0.82                                 & \multicolumn{1}{c|}{8.45}                                 & 23.41                                & 0.84                                 & 9.03                                 \\
RIED-Net                 & 10.27                                 & 0.44                                & \multicolumn{1}{c|}{82.05}                                & 24.51                                 & 0.84                                 & 5.60                                  & 3.79                                  & 0.30                                  & \multicolumn{1}{c|}{121.69}                               & 12.90                                 & 0.57                                 & 52.80                                 \\
ResViT                   & 27.34                                 & 0.86                                & \multicolumn{1}{c|}{4.83}                                 & 28.20                                  & 0.88                                 & 4.65                                 & \underline{25.91}                                 & \underline{0.89}                                 & \multicolumn{1}{c|}{\underline{6.53}}                                 & 25.18                                & 0.90                                  & 7.79                                 \\
\textbf{Ours}            & \textbf{29.25} & \textbf{0.90} & \multicolumn{1}{c|}{\textbf{3.96}} & \textbf{29.93} & \textbf{0.92} & \textbf{3.73} & \textbf{26.94} & \textbf{0.92} & \multicolumn{1}{c|}{\textbf{5.88}} & \textbf{26.40} & \textbf{0.93} & \textbf{6.77} \\ \hline
\end{tabular}}
\caption{Quantitative evaluation on IXI and BraTS2020 dataset. The best performance is marked in bold, while the second-best performance is underlined.}
\label{table:ixi}
\end{table*}
\begin{table*}[ht]
\adjustbox{width=\linewidth}{
\begin{tabular}{l|cccccc|cccccc}
\hline
\multirow{3}{*}{Method} & \multicolumn{6}{c|}{SynthRAD2023 Task1}                                                                                    & \multicolumn{6}{c}{SynthRAD2023 Task2}                                                                                     \\ \cline{2-13} 
                        & \multicolumn{3}{c|}{$\rm MRI\rightarrow CT$}                           & \multicolumn{3}{c|}{$\rm CT\rightarrow MRI$}      & \multicolumn{3}{c|}{$\rm CBCT\rightarrow CT$}                          & \multicolumn{3}{c}{$\rm CT\rightarrow CBCT$}      \\
                        & PSNR$\uparrow$ & SSIM$\uparrow$ & \multicolumn{1}{c|}{MAE$\downarrow$} & PSNR$\uparrow$ & SSIM$\uparrow$ & MAE$\downarrow$ & PSNR$\uparrow$ & SSIM$\uparrow$ & \multicolumn{1}{c|}{MAE$\downarrow$} & PSNR$\uparrow$ & SSIM$\uparrow$ & MAE$\downarrow$ \\ \hline
U-Net                   & 21.05          & 0.77           & \multicolumn{1}{c|}{14.42}           & 16.71          & 0.53           & 25.60            & 22.30           & 0.79           & \multicolumn{1}{c|}{14.54}           & 20.61          & 0.70           & 21.03           \\
CycleGAN                & 9.40            & 0.17           & \multicolumn{1}{c|}{69.07}           & 11.58          & 0.28           & 48.34           & 10.36          & 0.21           & \multicolumn{1}{c|}{64.19}           & 10.77          & 0.23           & 63.54           \\
Pix2Pix                 & 10.04          & 0.19           & \multicolumn{1}{c|}{65.80}            & 12.36          & 0.32           & 43.91           & 10.42          & 0.21           & \multicolumn{1}{c|}{63.45}           & 11.29          & 0.28           & 58.20            \\
UNIT                    & 9.65           & 0.17           & \multicolumn{1}{c|}{67.91}           & 12.50           & 0.31           & 42.17           & 10.29          & 0.20            & \multicolumn{1}{c|}{64.74}           & 10.94          & 0.24           & 62.60            \\
MUNIT                   & 10.12          & 0.20            & \multicolumn{1}{c|}{64.97}           & 12.49          & 0.31           & 42.65           & 10.46          & 0.21           & \multicolumn{1}{c|}{62.98}           & 11.24          & 0.27           & 57.88           \\
FUNIT                   & 8.93           & 0.04           & \multicolumn{1}{c|}{75.22}           & 8.47           & 0.11           & 74.19           & 10.28          & 0.46           & \multicolumn{1}{c|}{64.58}           & 9.88           & 0.44           & 70.57           \\
U-GAT-IT                  & 21.45          & 0.76           & \multicolumn{1}{c|}{13.07}           & 16.68          & 0.51           & 26.10            & 23.37          & 0.81           & \multicolumn{1}{c|}{12.45}           & 20.83          & 0.69           & 21.13           \\
CUT                     & 14.14          & 0.42           & \multicolumn{1}{c|}{44.87}           & 11.43          & 0.29           & 45.21           & 21.03          & 0.74           & \multicolumn{1}{c|}{18.27}           & 20.37          & 0.70            & 21.48           \\
LPTN                    & 16.79          & 0.56           & \multicolumn{1}{c|}{24.33}           & 13.37          & 0.34           & 36.74           & 22.15          & 0.79           & \multicolumn{1}{c|}{14.31}           & 21.28          & 0.71           & 18.59           \\
medSynth                & 15.11          & 0.34           & \multicolumn{1}{c|}{32.79}           & 15.81          & 0.40            & 29.52           & 20.52          & 0.71           & \multicolumn{1}{c|}{20.70}            & 19.90           & 0.68           & 21.62           \\
pGAN                    & 20.78          & 0.73           & \multicolumn{1}{c|}{15.04}           & 18.19          & 0.55           & 20.10            & 21.49          & 0.75           & \multicolumn{1}{c|}{14.93}           & 20.71          & 0.68           & 19.78           \\
RIED-Net                & 22.70           & \underline{0.80}      & \multicolumn{1}{c|}{10.84}           & 16.99          & 0.54           & 22.93           & 22.46          & \underline{0.82}     & \multicolumn{1}{c|}{12.56}           & 20.50           & \underline{0.72}     & 19.81           \\
ResViT                  & \underline{22.98}    & 0.79           & \multicolumn{1}{c|}{\textbf{10.39}}  & \underline{18.88}    & \underline{0.58}     & \underline{17.59}     & \underline{24.15}    & \underline{0.82}     & \multicolumn{1}{c|}{\textbf{9.80}}    & \underline{22.87}    & \underline{0.72}     & \underline{15.58}     \\
\textbf{Ours}                    & \textbf{23.33} & \textbf{0.83}  & \multicolumn{1}{c|}{\underline{10.63}}     & \textbf{19.99} & \textbf{0.66}  & \textbf{15.91}  & \textbf{24.67} & \textbf{0.85}  & \multicolumn{1}{c|}{\underline{9.83}}      & \textbf{23.95} & \textbf{0.79}  & \textbf{13.35}  \\ \hline
\end{tabular}}
\caption{Quantitative evaluation on SynthRAD2023 dataset. The best performance is marked in bold, while the second-best performance is underlined.}
\label{table:syn}
\end{table*}

\begin{table*}[ht]
\centering
\adjustbox{width=\linewidth}{
\begin{tabular}{ccc|cccc}
\hline
\multicolumn{3}{l|}{}                                                                               & Ours w/o PEB & Ours w/o PFB & Ours w/o Transformer & Ours Full \\ \hline
\multicolumn{1}{c|}{\multirow{6}{*}{IXI}}                & \multirow{3}{*}{$\rm T_{1}\rightarrow T_{2}$} & PSNR$\uparrow$  &   25.65    &   26.06   &   27.17      &   \textbf{29.25}        \\
\multicolumn{1}{c|}{}                                    &                        & SSIM$\uparrow$  &   0.82   &   0.83    &     0.87    &    \textbf{0.90}       \\
\multicolumn{1}{c|}{}                                    &                        & MAE$\downarrow$ &    6.37   &   5.71    &     5.05    &    \textbf{3.96}       \\ \cline{2-7} 
\multicolumn{1}{c|}{}                                    & \multirow{3}{*}{$\rm T_{2}\rightarrow T_{1}$} & PSNR$\uparrow$  &    26.99   &    26.66   &   27.82      &   \textbf{29.93}        \\
\multicolumn{1}{c|}{}                                    &                        & SSIM$\uparrow$  &     0.85  &    0.81   &     0.89    &     \textbf{0.92}      \\
\multicolumn{1}{c|}{}                                    &                        & MAE$\downarrow$ &  5.42     &    6.25   &     4.87    &     \textbf{3.73}      \\ \hline
\multicolumn{1}{c|}{\multirow{6}{*}{BraTS2020}}          & \multirow{3}{*}{$\rm T_{1}\rightarrow T_{2}$} & PSNR$\uparrow$  &   24.55    &   24.92    &    25.86     &   \textbf{26.94}        \\
\multicolumn{1}{c|}{}                                    &                        & SSIM$\uparrow$  &    0.88   &   0.88    &    0.91     &   \textbf{0.92}        \\
\multicolumn{1}{c|}{}                                    &                        & MAE$\downarrow$ &   7.57    &    7.35   &   6.58      &    \textbf{5.88}       \\ \cline{2-7} 
\multicolumn{1}{c|}{}                                    & \multirow{3}{*}{$\rm T_{2}\rightarrow T_{1}$} & PSNR$\uparrow$  &   25.28    &   25.44    &    25.58     &   \textbf{26.40}        \\
\multicolumn{1}{c|}{}                                    &                        & SSIM$\uparrow$  &   0.90    &   0.90    &   0.92      &   \textbf{0.93}        \\
\multicolumn{1}{c|}{}                                    &                        & MAE$\downarrow$ &   7.45    &    7.57   &    7.11     &   \textbf{6.77}        \\ \hline
\multicolumn{1}{c|}{\multirow{6}{*}{SynthRAD2023 Task1}} & \multirow{3}{*}{$\rm MRI\rightarrow CT$} & PSNR$\uparrow$  &    22.19   &   22.31    &    22.04     &    \textbf{23.33}       \\
\multicolumn{1}{c|}{}                                    &                        & SSIM$\uparrow$  &   0.78    &    0.78   &    0.78     &   \textbf{0.83}        \\
\multicolumn{1}{c|}{}                                    &                        & MAE$\downarrow$ &   12.39    &    12.28   &    12.39     &     \textbf{10.63}      \\ \cline{2-7} 
\multicolumn{1}{c|}{}                                    & \multirow{3}{*}{$\rm CT\rightarrow MRI$} & PSNR$\uparrow$  &   19.08    &  19.10     &  19.06       & \textbf{19.99}          \\
\multicolumn{1}{c|}{}                                    &                        & SSIM$\uparrow$  &   0.60    &    0.59   &    0.61     &    \textbf{0.66}       \\
\multicolumn{1}{c|}{}                                    &                        & MAE$\downarrow$ &   18.54    &   19.02    &    18.75     &   \textbf{15.91}        \\ \hline
\multicolumn{1}{c|}{\multirow{6}{*}{SynthRAD2023 Task2}} & \multirow{3}{*}{$\rm CBCT\rightarrow CT$} & PSNR$\uparrow$  &       23.57 & 23.91 & 23.79 & \textbf{24.67}          \\
\multicolumn{1}{c|}{}                                    &                        & SSIM$\uparrow$  &       0.82  & 0.82  & 0.84  & \textbf{0.85}           \\
\multicolumn{1}{c|}{}                                    &                        & MAE$\downarrow$ &       11.80  & 11.03 & 11.28 & \textbf{9.83}           \\ \cline{2-7} 
\multicolumn{1}{c|}{}                                    & \multirow{3}{*}{$\rm CT\rightarrow CBCT$} & PSNR$\uparrow$  &       22.28 & 22.65 & 22.23 & \textbf{23.95}           \\
\multicolumn{1}{c|}{}                                    &                        & SSIM$\uparrow$  &       0.76  & 0.75  & 0.73  & \textbf{0.79}           \\
\multicolumn{1}{c|}{}                                    &                        & MAE$\downarrow$ &       16.62 & 15.98 & 16.75 & \textbf{13.35}           \\ \hline
\multicolumn{1}{c|}{\multirow{6}{*}{ADNI}}               & \multirow{3}{*}{$\rm MRI\rightarrow PET$} & PSNR$\uparrow$  &       21.20  & 21.23 & 21.20  & \textbf{24.43}           \\
\multicolumn{1}{c|}{}                                    &                        & SSIM$\uparrow$  &       0.71  & 0.71  & 0.71  & \textbf{0.84}           \\
\multicolumn{1}{c|}{}                                    &                        & MAE$\downarrow$ &       14.86 & 14.86 & 14.81 & \textbf{9.73}           \\ \cline{2-7} 
\multicolumn{1}{c|}{}                                    & \multirow{3}{*}{$\rm PET\rightarrow MRI$} & PSNR$\uparrow$  &       18.41 & 18.28 & 18.40  & \textbf{21.00}           \\
\multicolumn{1}{c|}{}                                    &                        & SSIM$\uparrow$  &       0.66  & 0.64  & 0.62  & \textbf{0.79}           \\
\multicolumn{1}{c|}{}                                    &                        & MAE$\downarrow$ &       17.29 & 17.73 & 18.01 & \textbf{12.39}           \\ \hline
\end{tabular}
}
\caption{Ablation study on PEB, PFB, and Transformer blocks.}
\label{tabel:ab}
\end{table*}

\subsubsection{Evaluation Metrics}
For the evaluation metrics, we employ three widely-used metrics: Peak Signal-to-Noise Ratio (PSNR), Structural Similarity (SSIM), and Mean Absolute Error (MAE). PSNR and SSIM are commonly employed in image translation evaluations and various low-level computer vision tasks. On the other hand, MAE provides a more general and conservative measurement of pixel misalignment by calculating the mean of absolute errors.

\subsubsection{Implementation Details}
The model is implemented using PyTorch and trained on the NVIDIA RTX 2080Ti GPU. We utilize the Adam optimizer with default parameters for training. When training the model, we set the batch size to 1 and the learning rate to $1e-4$. Furthermore, we apply several augmentation techniques to the training images, such as random cropping, resizing, rotation, flipping, and mixup.

\subsection{Comparisons with State-of-the-Arts}

In this section, we conduct extensive experiments, comparing our proposed method with a total of thirteen state-of-the-art methods. These include general-purpose image translation models: CycleGAN~\cite{zhu2017unpaired}, Pix2Pix~\cite{isola2017image}, UNIT~\cite{liu2017unsupervised}, MUNIT~\cite{huang2018multimodal}, FUNIT~\cite{liu2019few}, U-GAT-IT~\cite{kim2019u}, CUT~\cite{park2020contrastive}, LPTN~\cite{liang2021high}. Moreover, we incorporate medical image translation models: medSynth~\cite{nie2017medical}, pGAN~\cite{dar2019image}, RIED-Net~\cite{gao2019deep}, ResViT~\cite{dalmaz2022resvit}, and include U-Net~\cite{ronneberger2015u} as a baseline benchmark model.


As shown in Table~\ref{table:ixi}, our proposed method demonstrates superior performance compared to all other methods on the IXI and BraTS2020 datasets. It outperforms them in terms of various evaluation metrics, showcasing its effectiveness in cross-modal medical image translation tasks. On these two datasets, we surpassed the second-place performance by an average margin of 1 dB in terms of PSNR.

Although in Table~\ref{table:syn} our method did not achieve the lowest MAE compared to ResViT, it outperformed all other methods in terms of other evaluation metrics. This highlights the overall superiority of our approach in terms of translation quality and generalization capability. Additionally, our approach requires only single-stage training, surpassing all other methods in terms of convenience.

\begin{figure*}[!ht]
    \begin{minipage}[b]{1.0\linewidth}
        \begin{minipage}[b]{0.12\linewidth}
            \centering
            \centerline{\includegraphics[width=\linewidth]{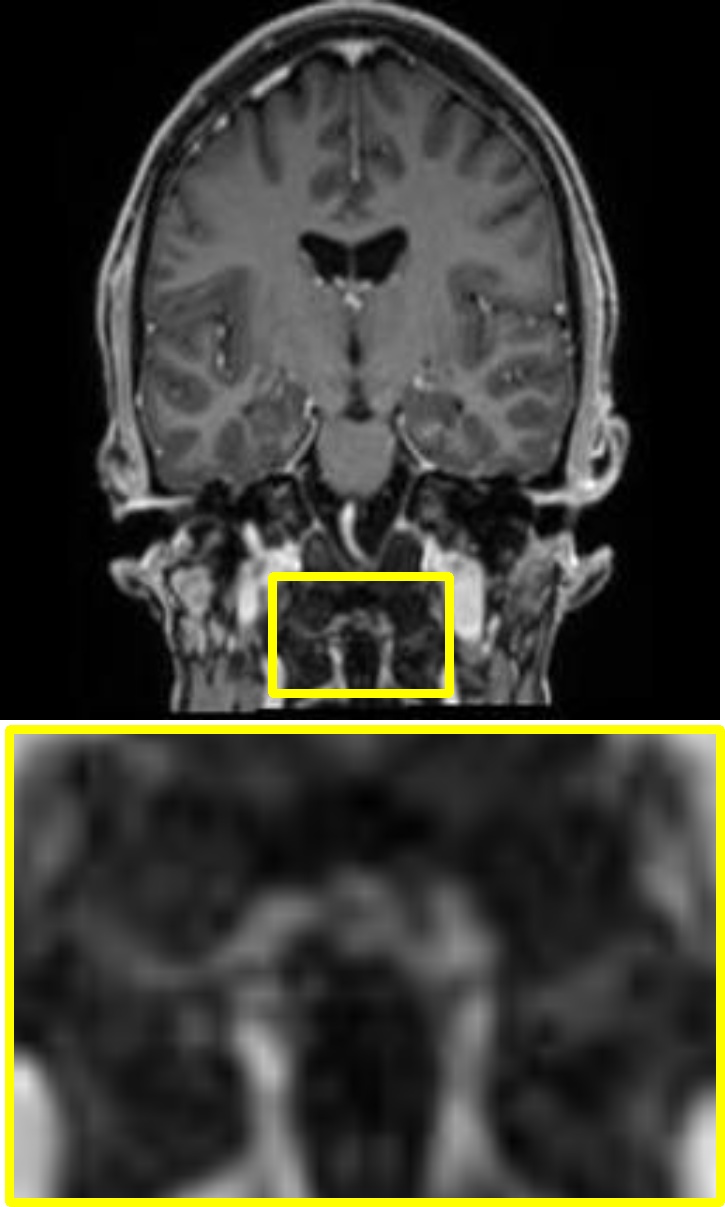}}
        \end{minipage}
        \begin{minipage}[b]{0.12\linewidth}
            \centering
            \centerline{\includegraphics[width=\linewidth]{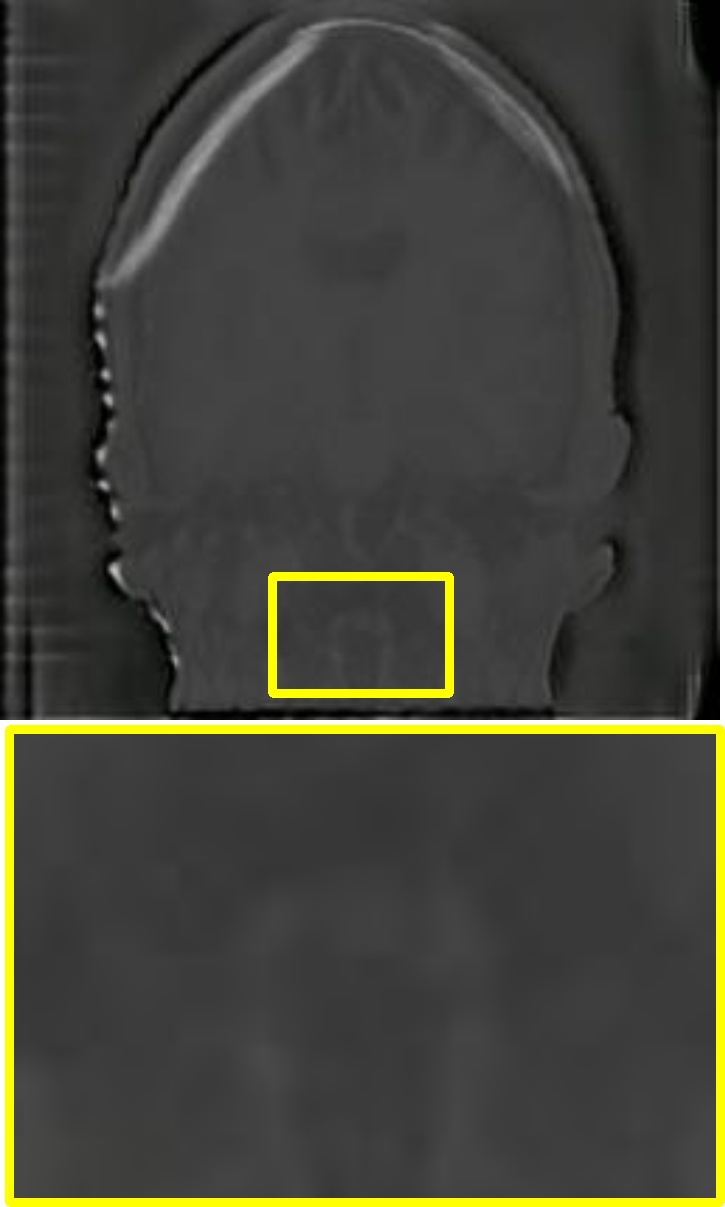}}
        \end{minipage}
        \begin{minipage}[b]{0.12\linewidth}
            \centering
            \centerline{\includegraphics[width=\linewidth]{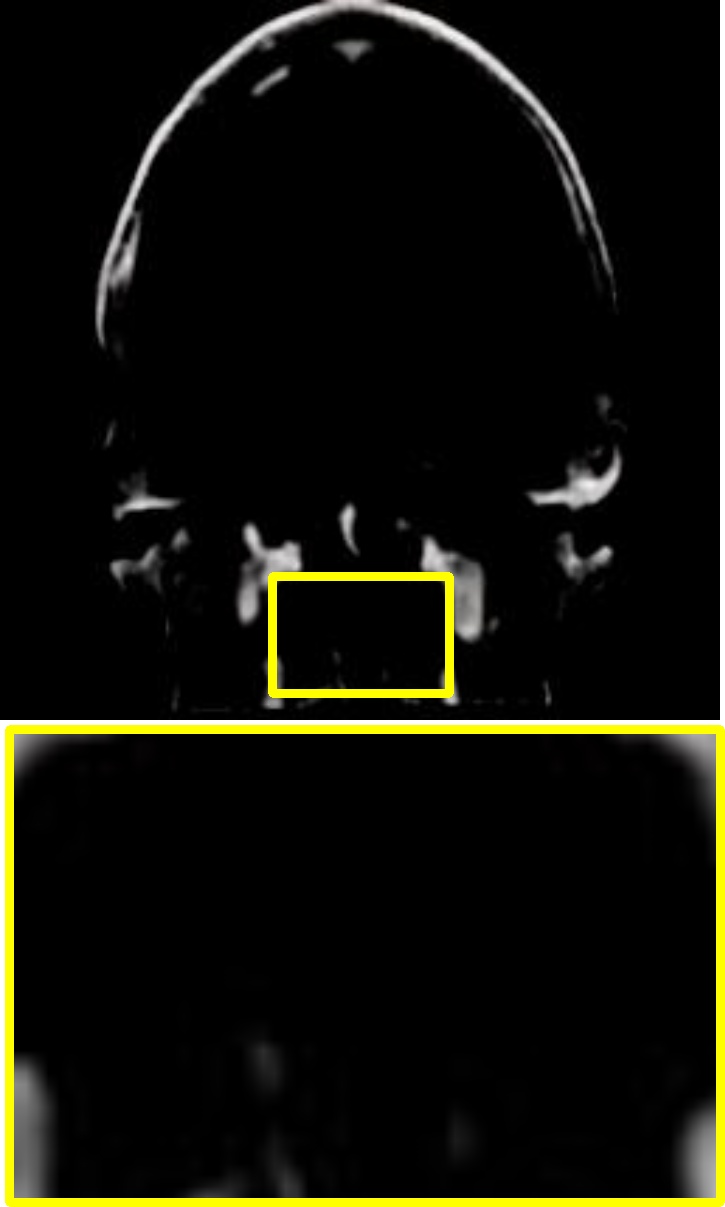}}
        \end{minipage}
        \begin{minipage}[b]{0.12\linewidth}
            \centering
            \centerline{\includegraphics[width=\linewidth]{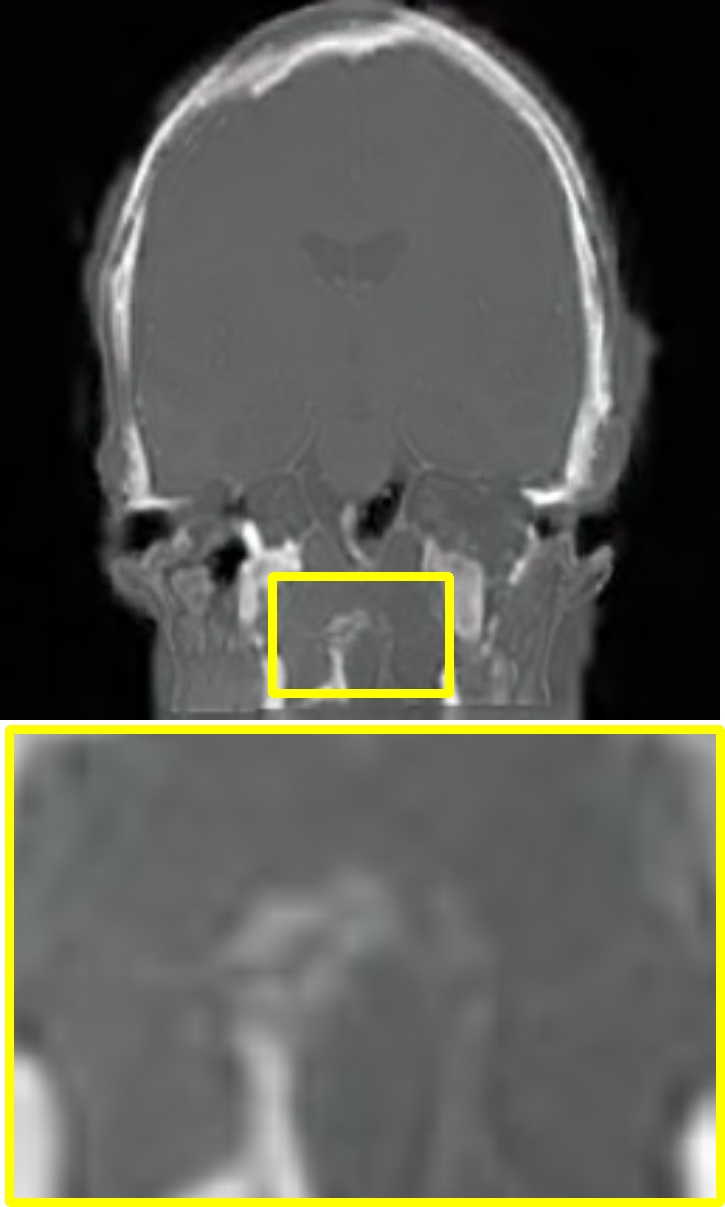}}
        \end{minipage} 
        \begin{minipage}[b]{0.12\linewidth}
            \centering
            \centerline{\includegraphics[width=\linewidth]{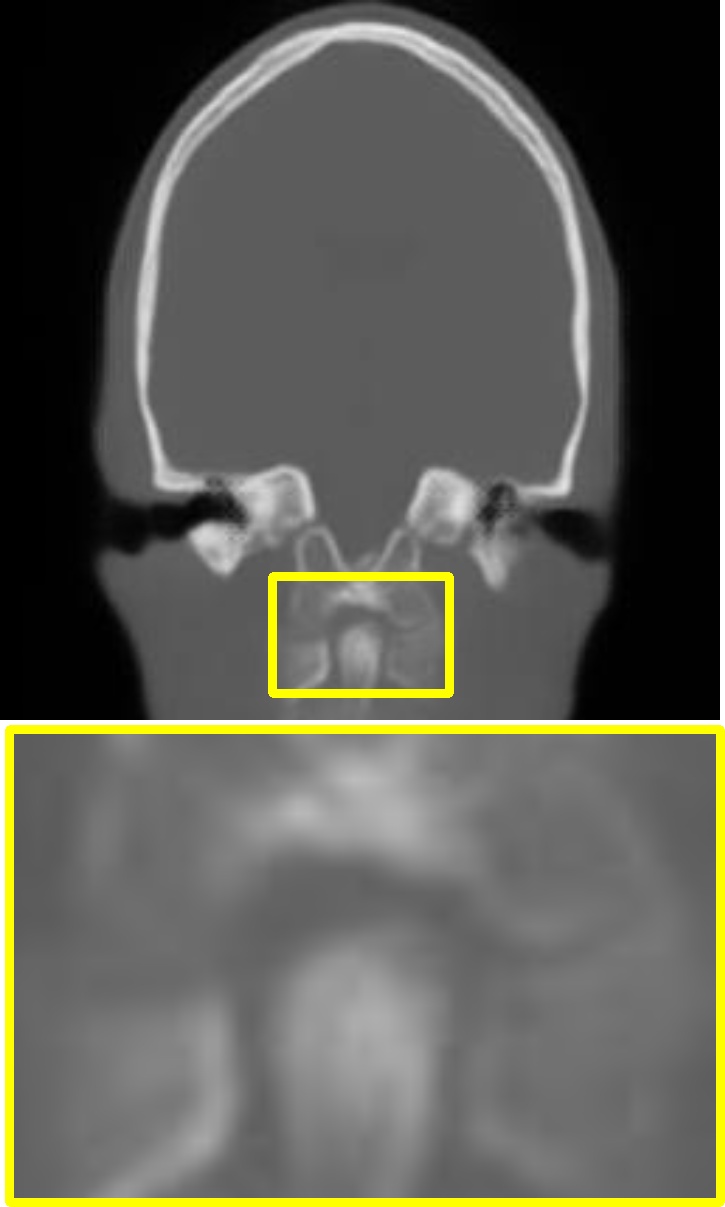}}
        \end{minipage}  
        \begin{minipage}[b]{0.12\linewidth}
            \centering
            \centerline{\includegraphics[width=\linewidth]{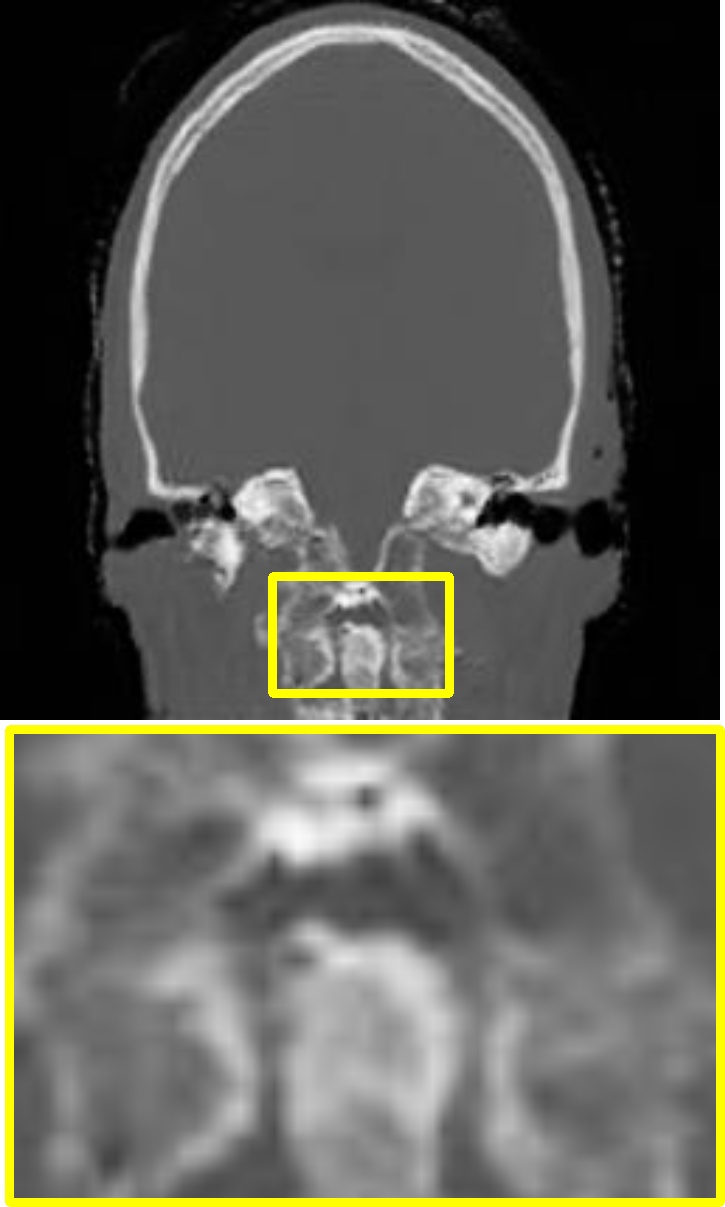}}
        \end{minipage}  
        \begin{minipage}[b]{0.12\linewidth}
            \centering
            \centerline{\includegraphics[width=\linewidth]{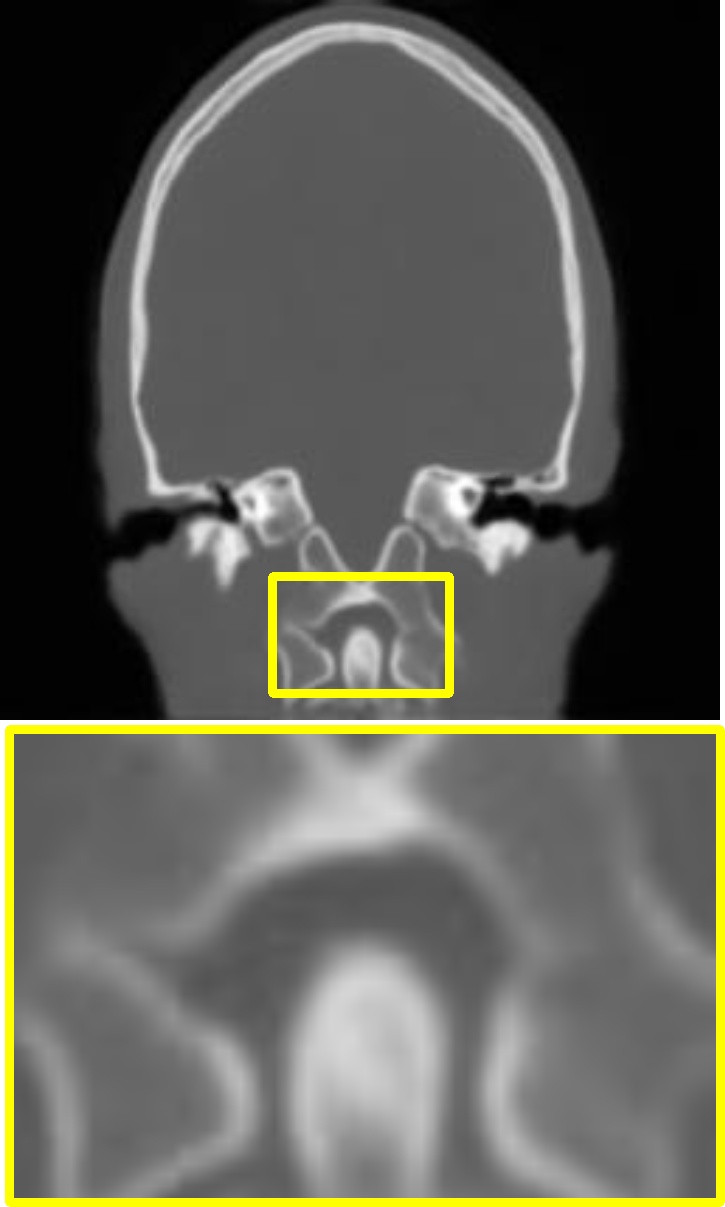}}
        \end{minipage}
        \begin{minipage}[b]{0.12\linewidth}
            \centering
            \centerline{\includegraphics[width=\linewidth]{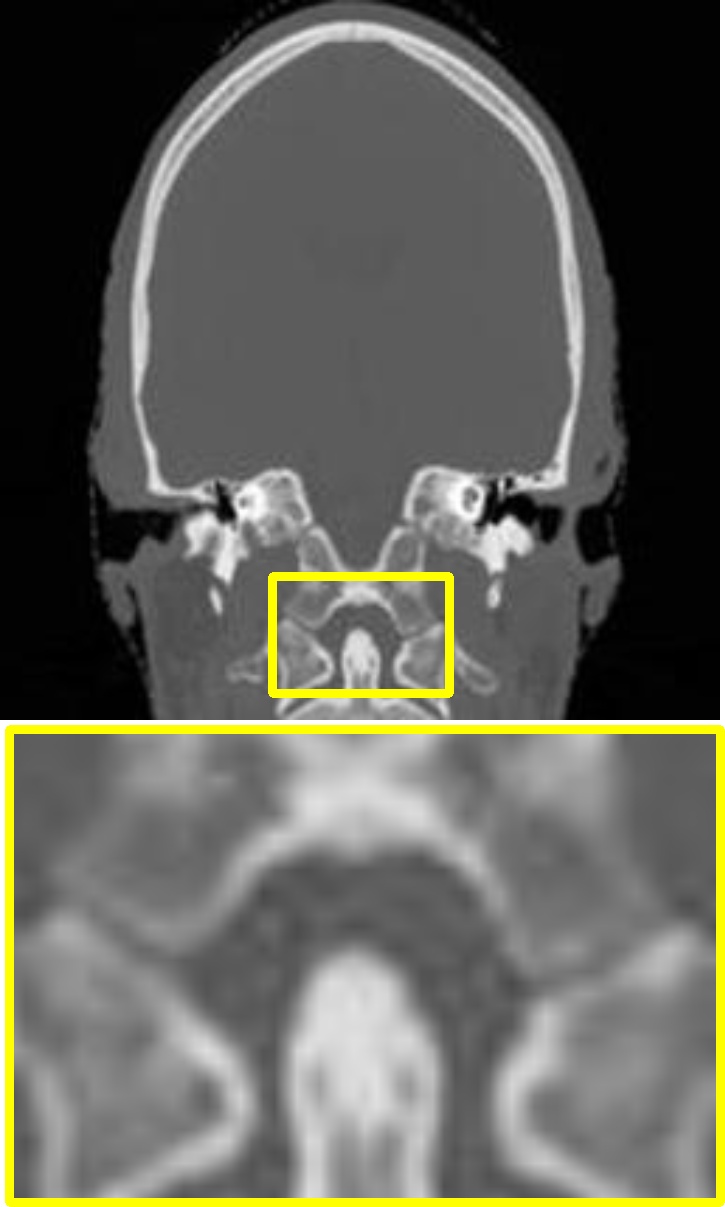}}
        \end{minipage}
    \end{minipage}
    
    \begin{minipage}[b]{1.0\linewidth}  
        \begin{minipage}[b]{0.12\linewidth}
            \centering
            \centerline{\includegraphics[width=\linewidth]{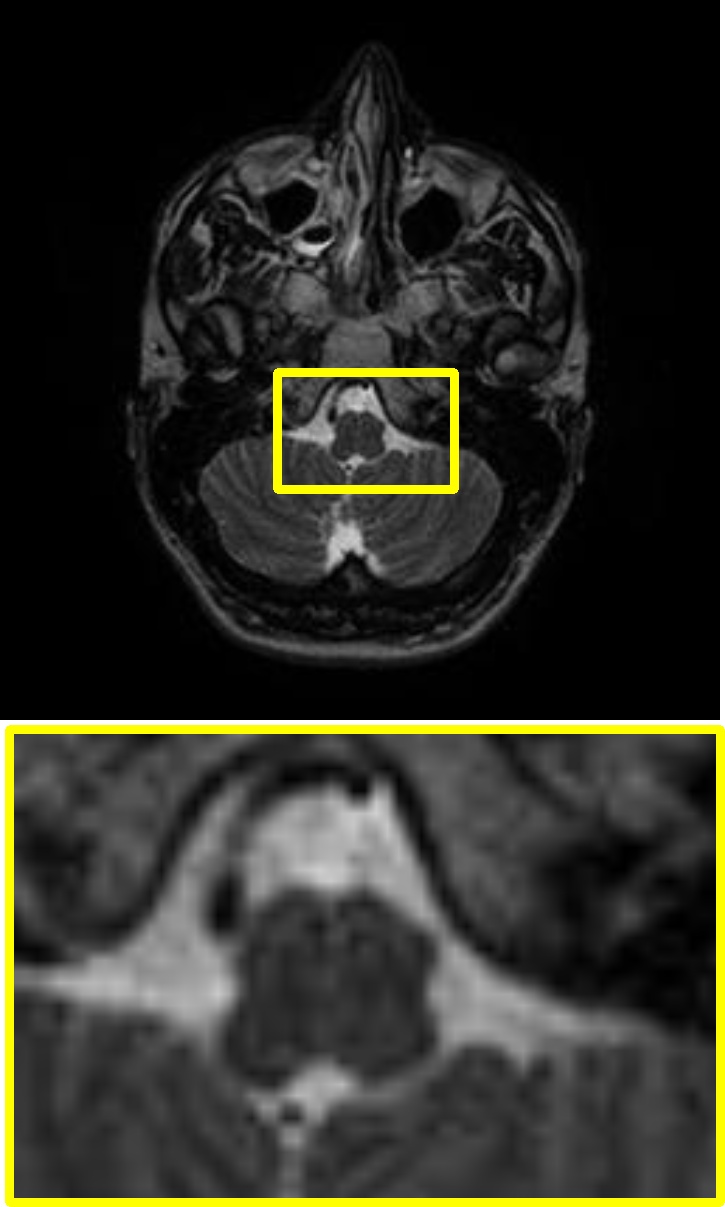}}
        \end{minipage}
        \begin{minipage}[b]{0.12\linewidth}
            \centering
            \centerline{\includegraphics[width=\linewidth]{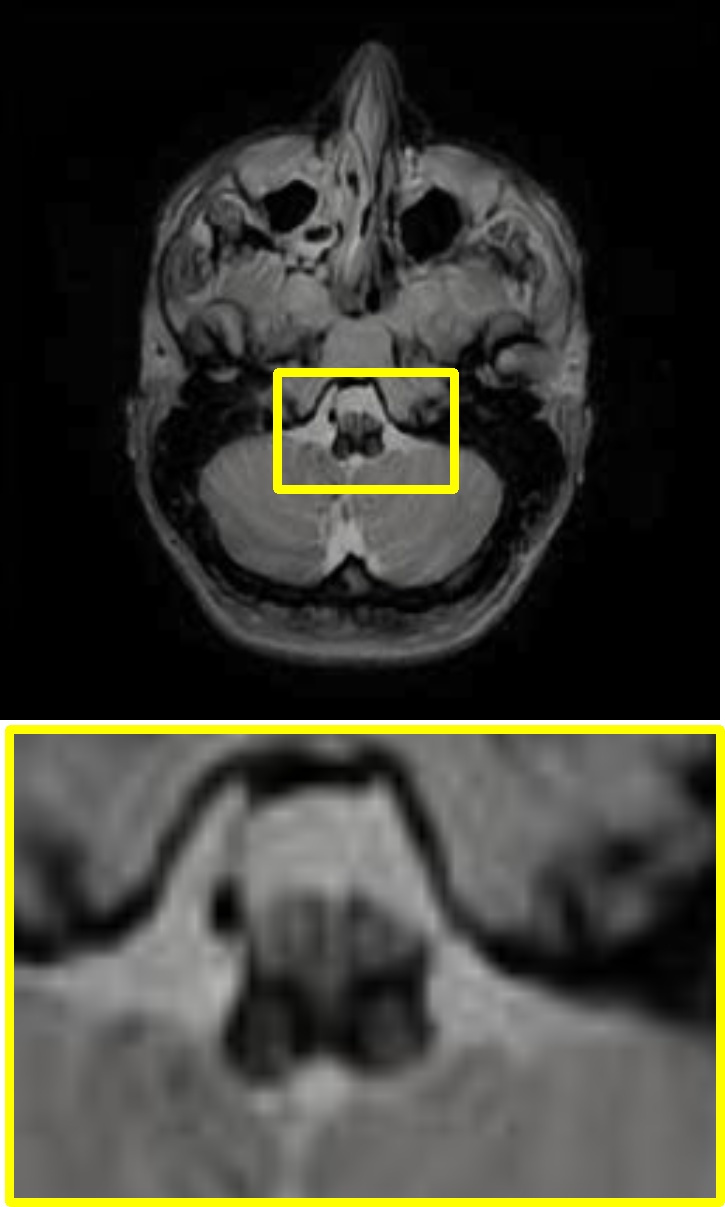}}
        \end{minipage}
        \begin{minipage}[b]{0.12\linewidth}
            \centering
            \centerline{\includegraphics[width=\linewidth]{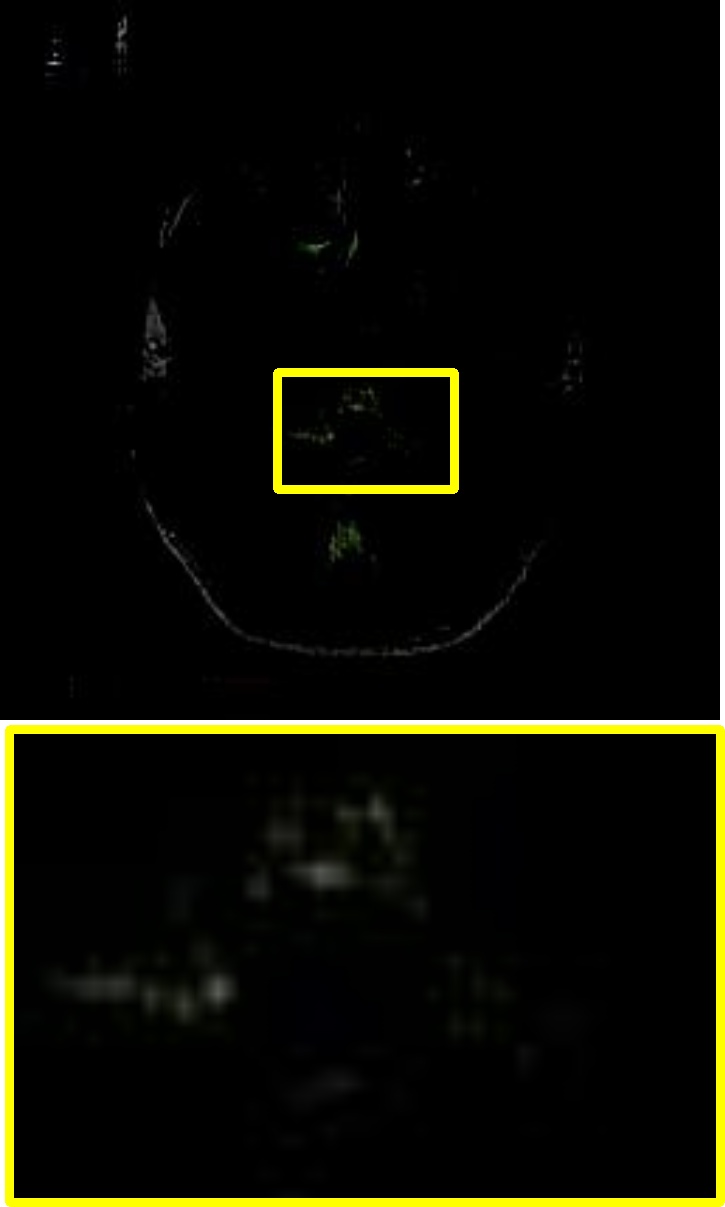}}
        \end{minipage}
        \begin{minipage}[b]{0.12\linewidth}
            \centering
            \centerline{\includegraphics[width=\linewidth]{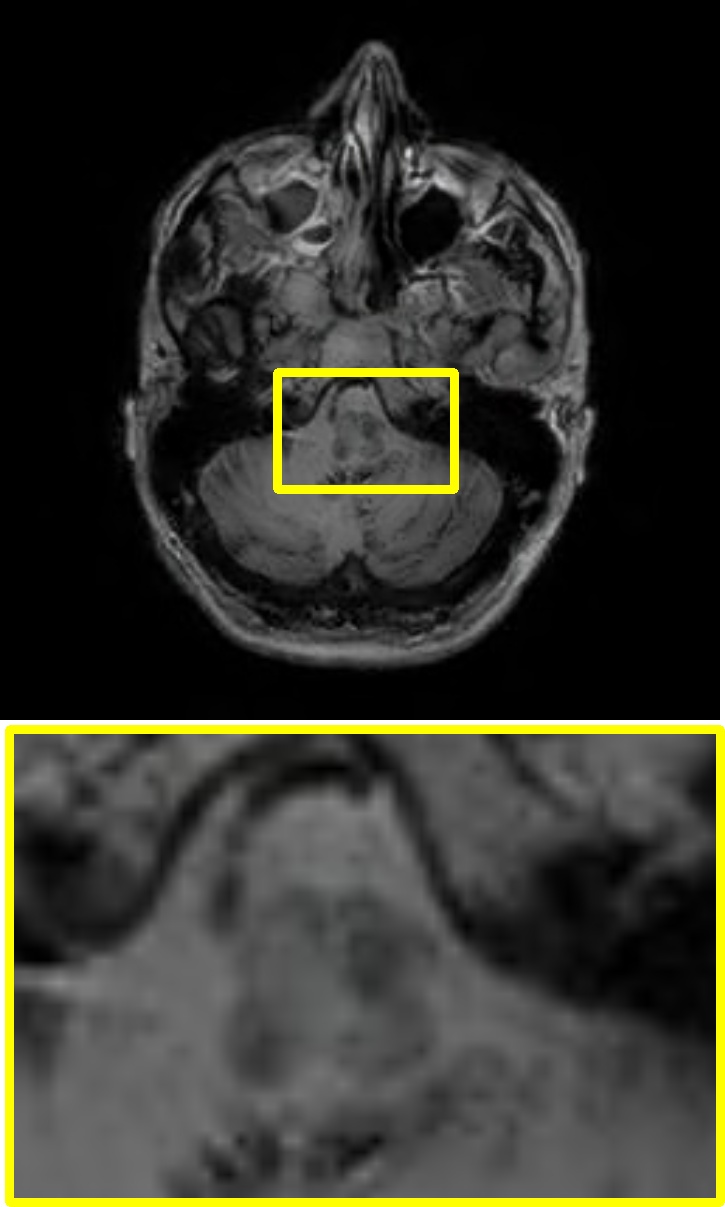}}
        \end{minipage}
        \begin{minipage}[b]{0.12\linewidth}
            \centering
            \centerline{\includegraphics[width=\linewidth]{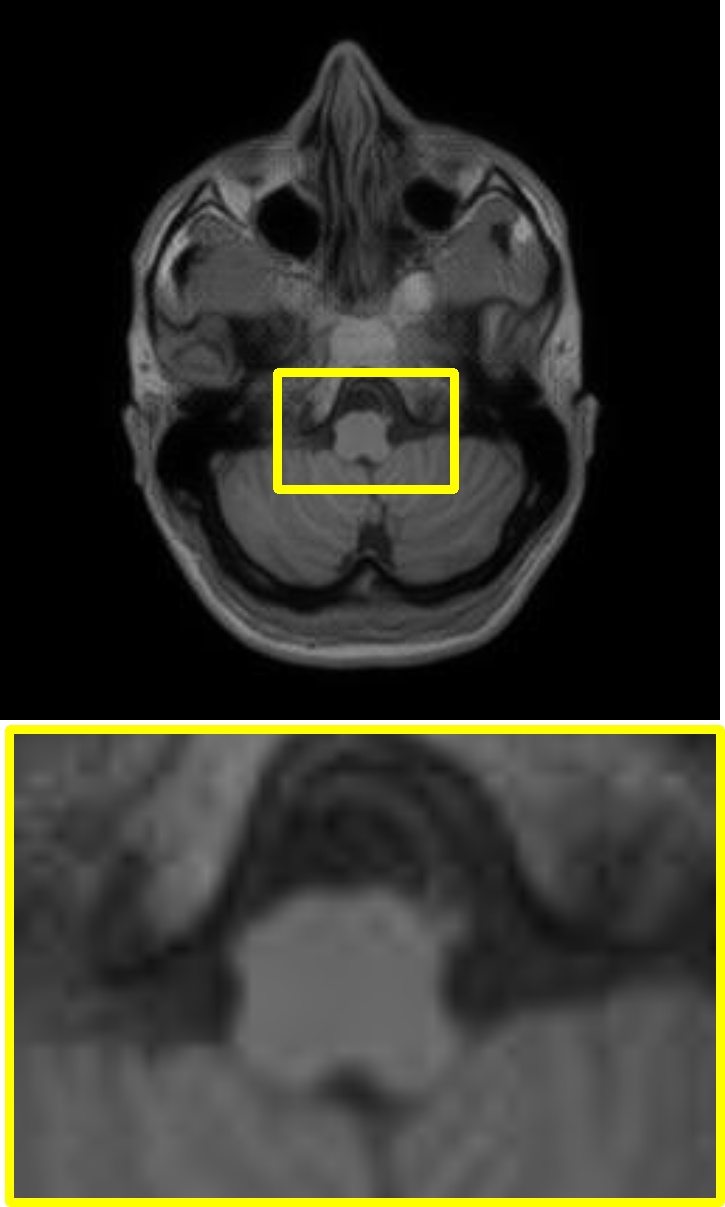}}
        \end{minipage}
        \begin{minipage}[b]{0.12\linewidth}
            \centering
            \centerline{\includegraphics[width=\linewidth]{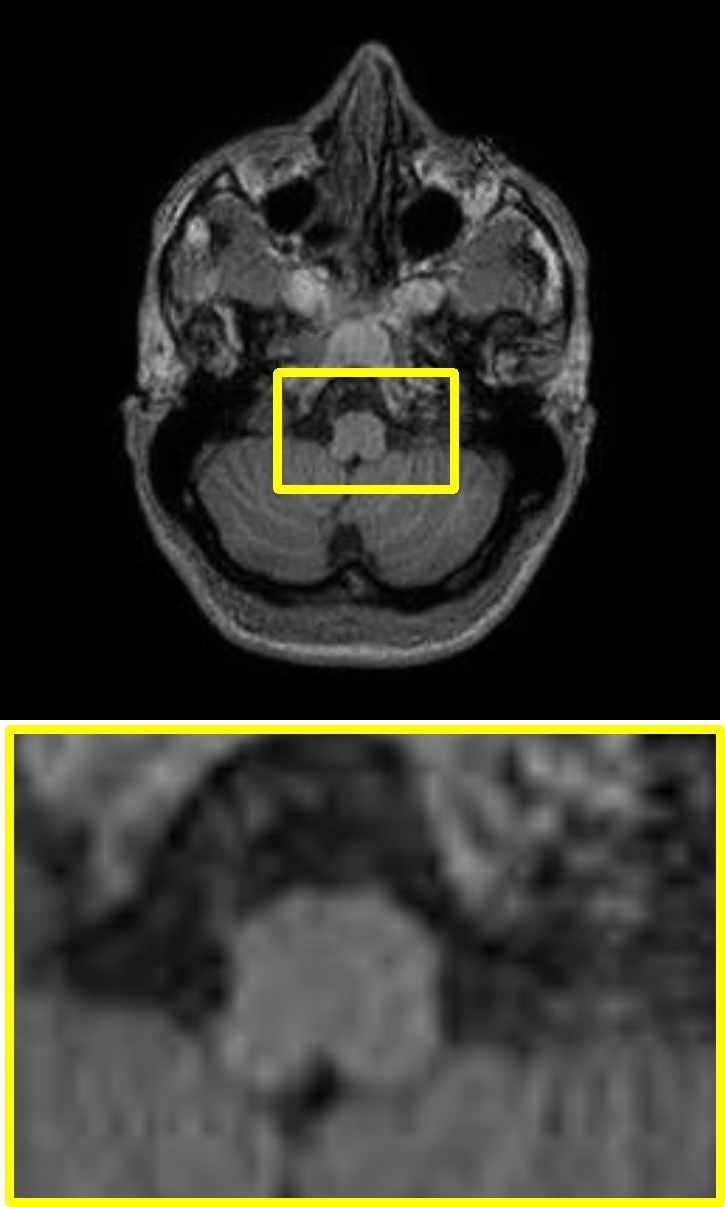}}
        \end{minipage}
        \begin{minipage}[b]{0.12\linewidth}
            \centering
            \centerline{\includegraphics[width=\linewidth]{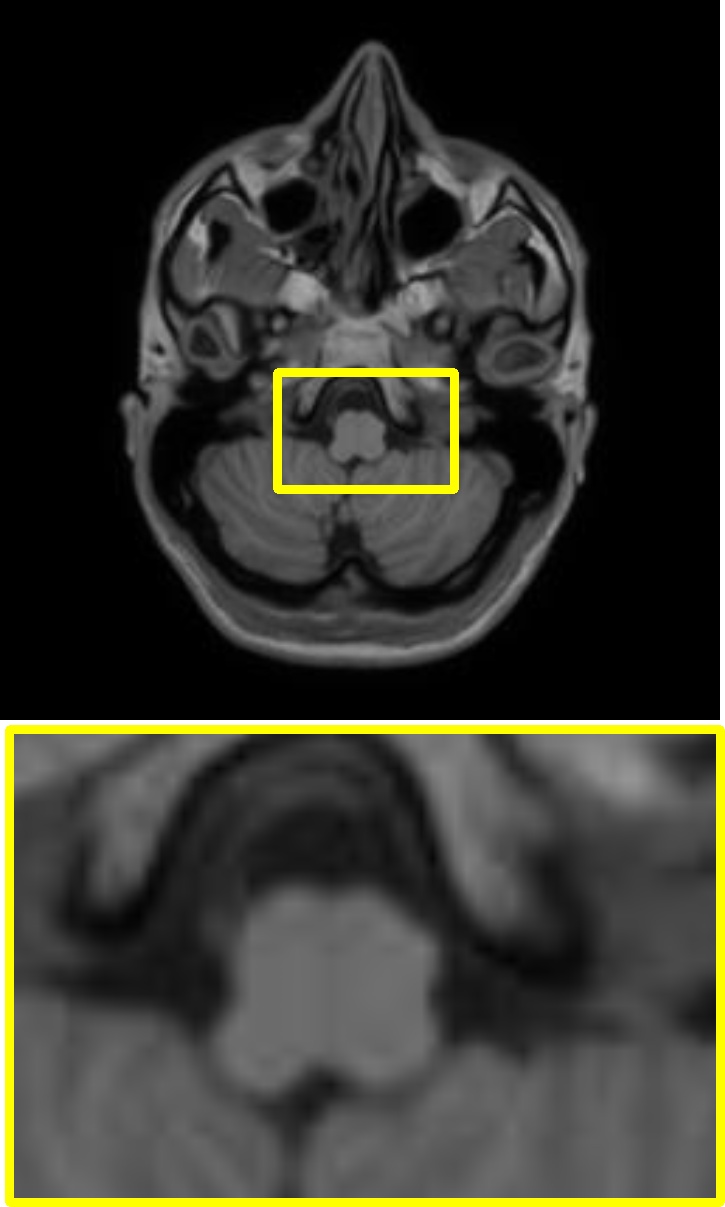}}
        \end{minipage}
        \begin{minipage}[b]{0.12\linewidth}
            \centering
            \centerline{\includegraphics[width=\linewidth]{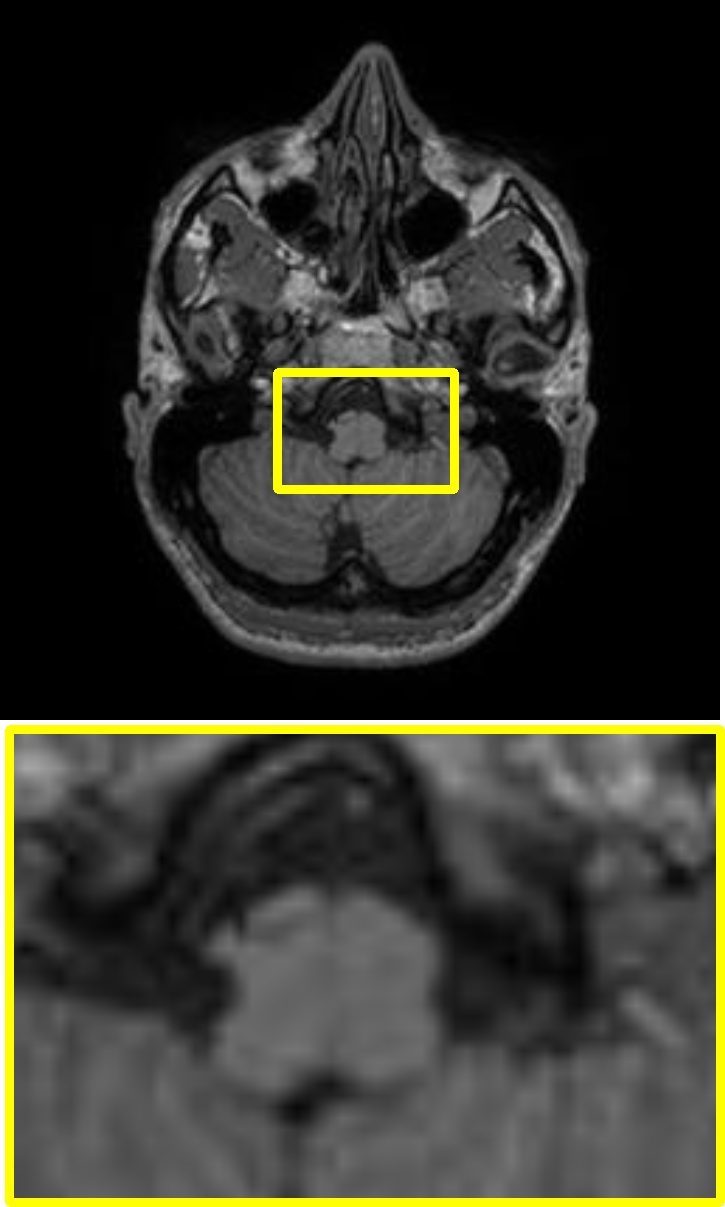}}
        \end{minipage}
    \end{minipage}

    \begin{minipage}[b]{1.0\linewidth}  
        \begin{minipage}[b]{0.12\linewidth}
            \centering
            \centerline{\includegraphics[width=\linewidth]{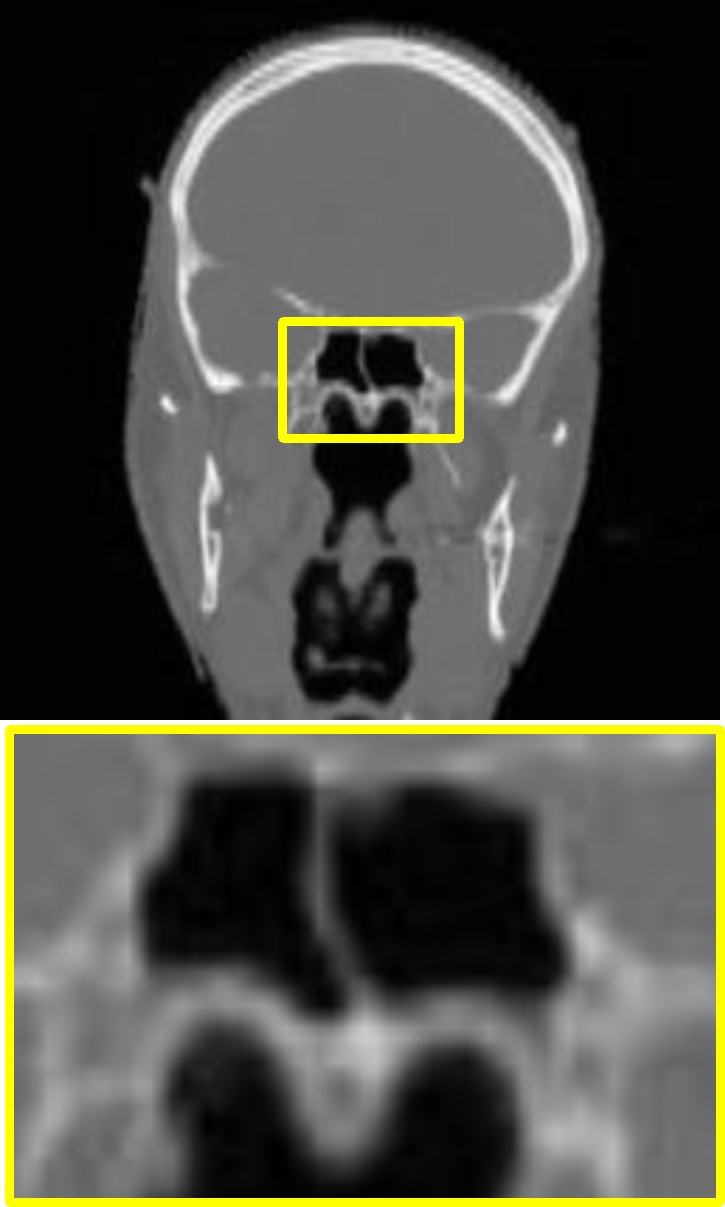}}
        \end{minipage}
        \begin{minipage}[b]{0.12\linewidth}
            \centering
            \centerline{\includegraphics[width=\linewidth]{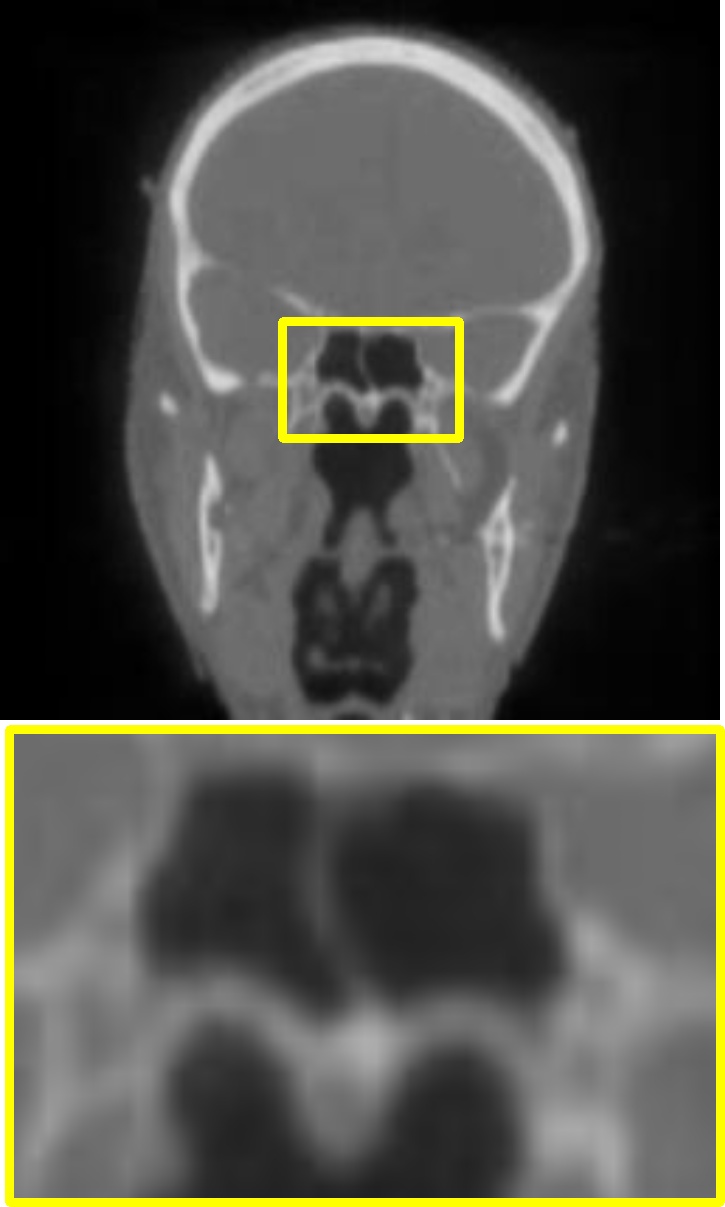}}
        \end{minipage}
        \begin{minipage}[b]{0.12\linewidth}
            \centering
            \centerline{\includegraphics[width=\linewidth]{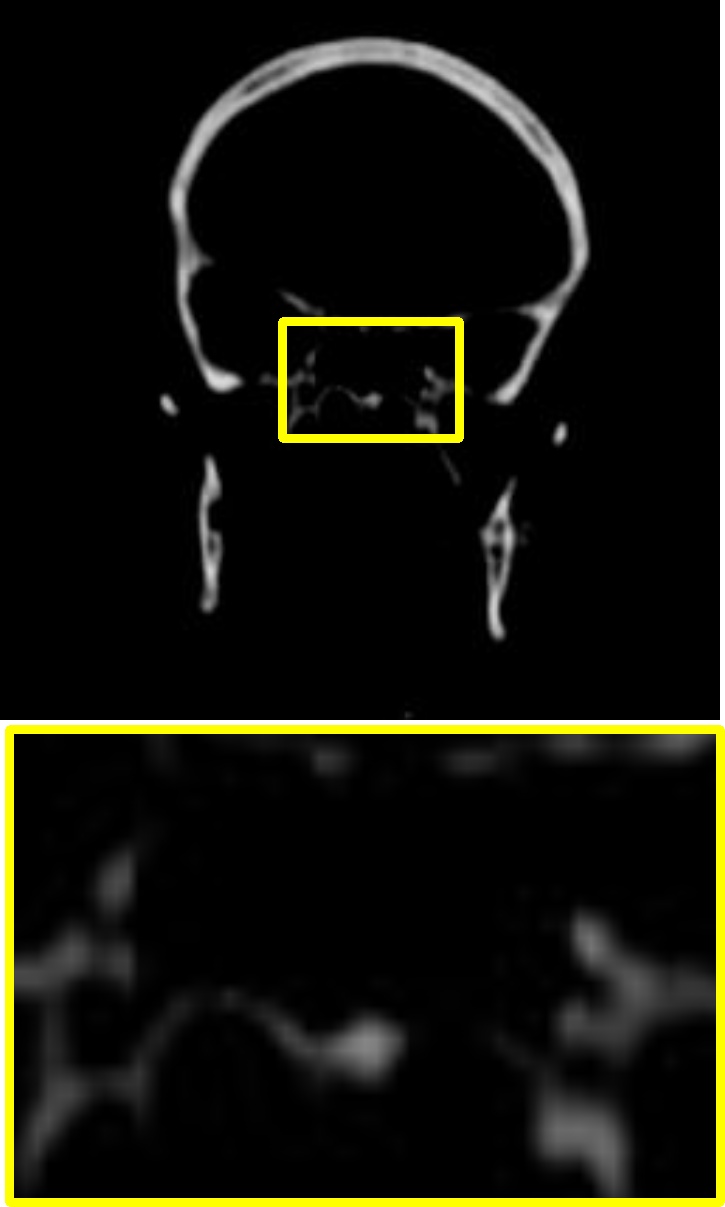}}
        \end{minipage}
        \begin{minipage}[b]{0.12\linewidth}
            \centering
            \centerline{\includegraphics[width=\linewidth]{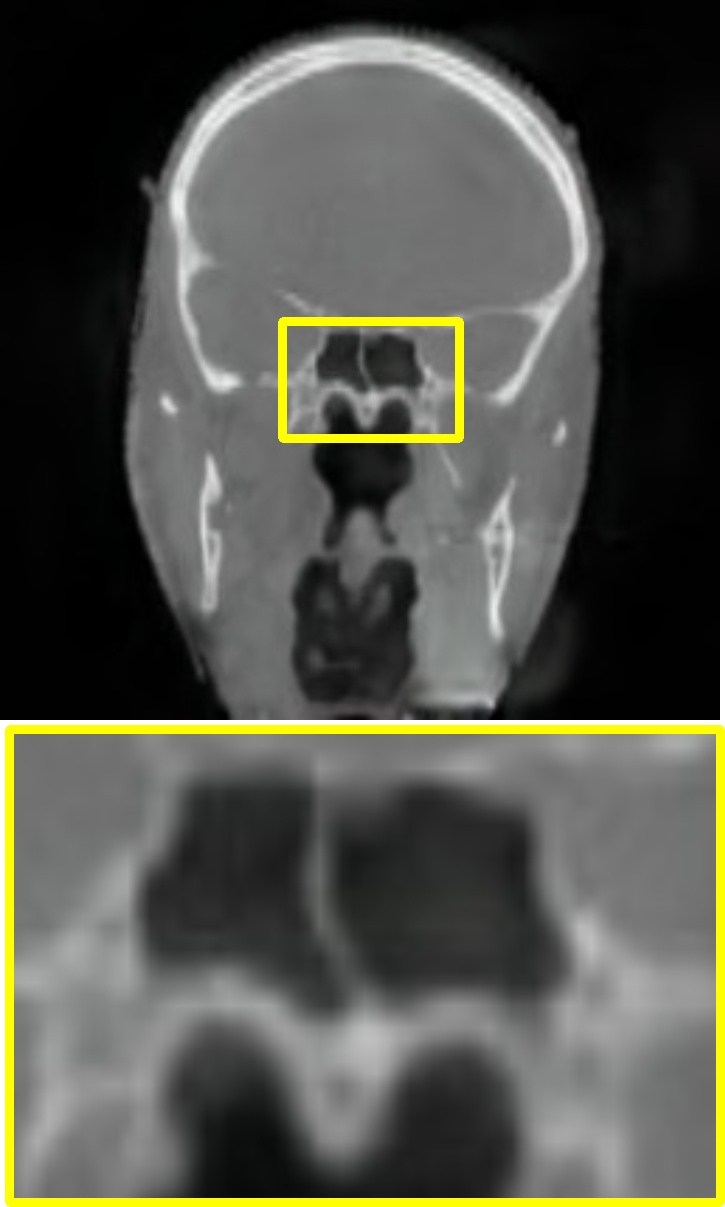}}
        \end{minipage}
        \begin{minipage}[b]{0.12\linewidth}
            \centering
            \centerline{\includegraphics[width=\linewidth]{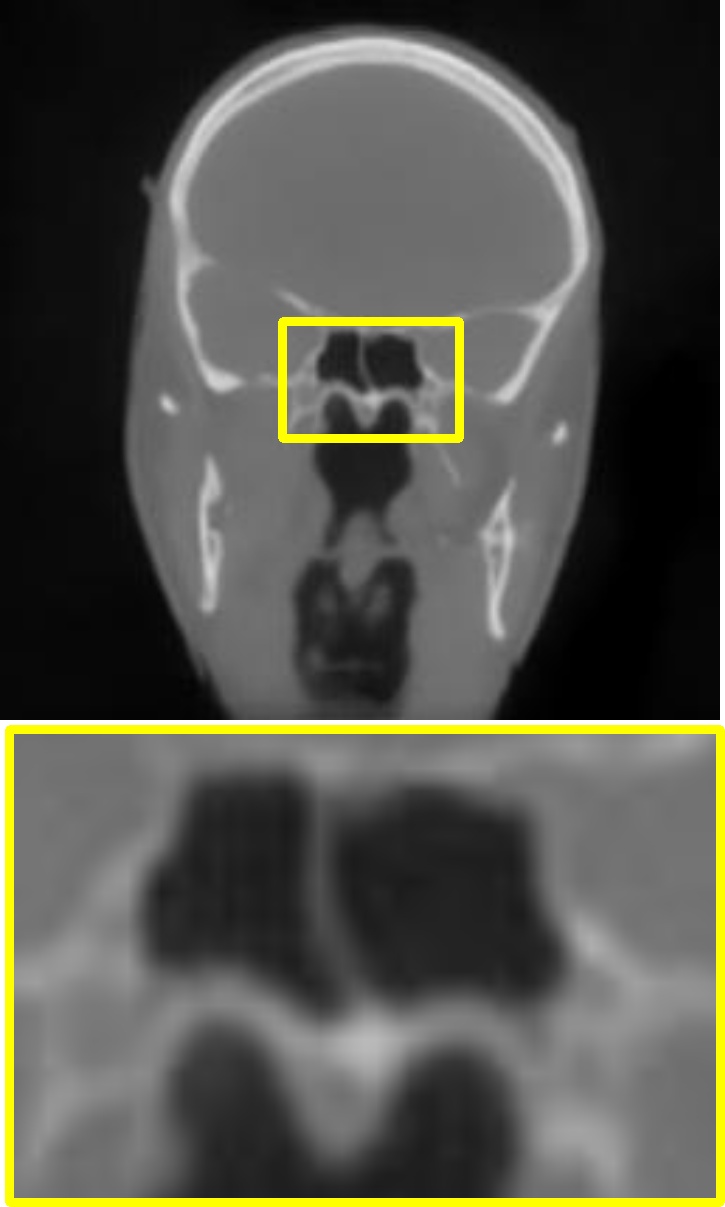}}
        \end{minipage}
        \begin{minipage}[b]{0.12\linewidth}
            \centering
            \centerline{\includegraphics[width=\linewidth]{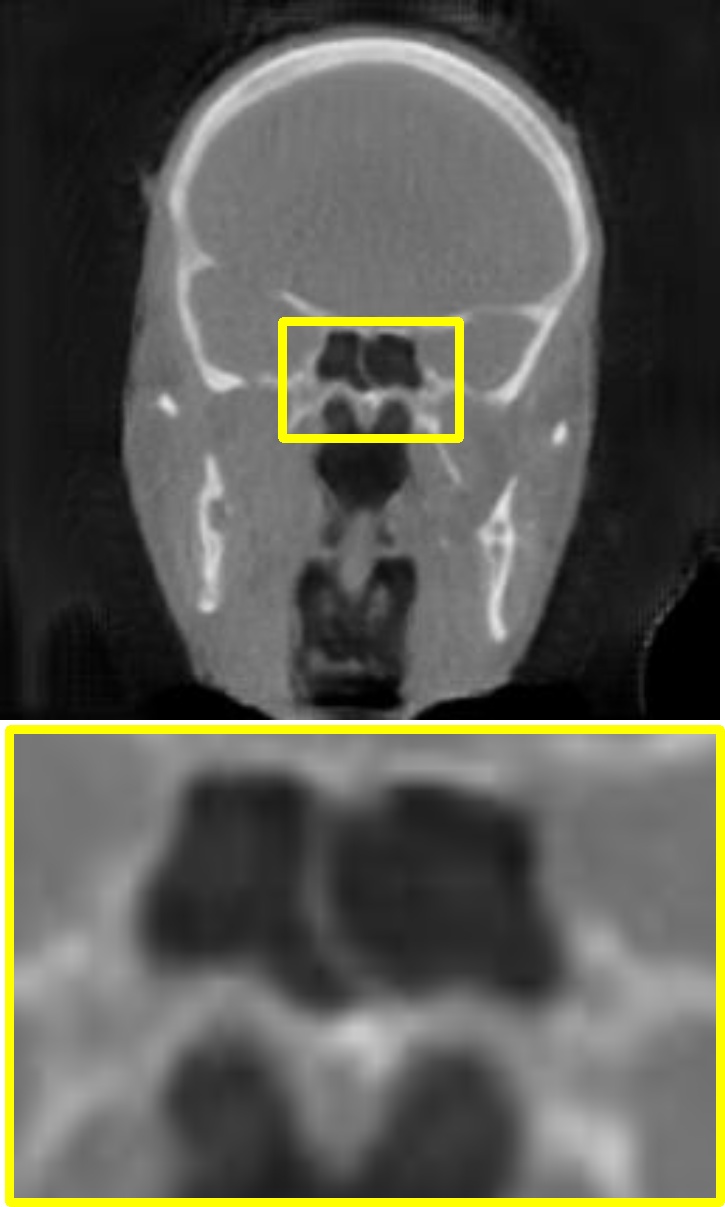}}
        \end{minipage}
        \begin{minipage}[b]{0.12\linewidth}
            \centering
            \centerline{\includegraphics[width=\linewidth]{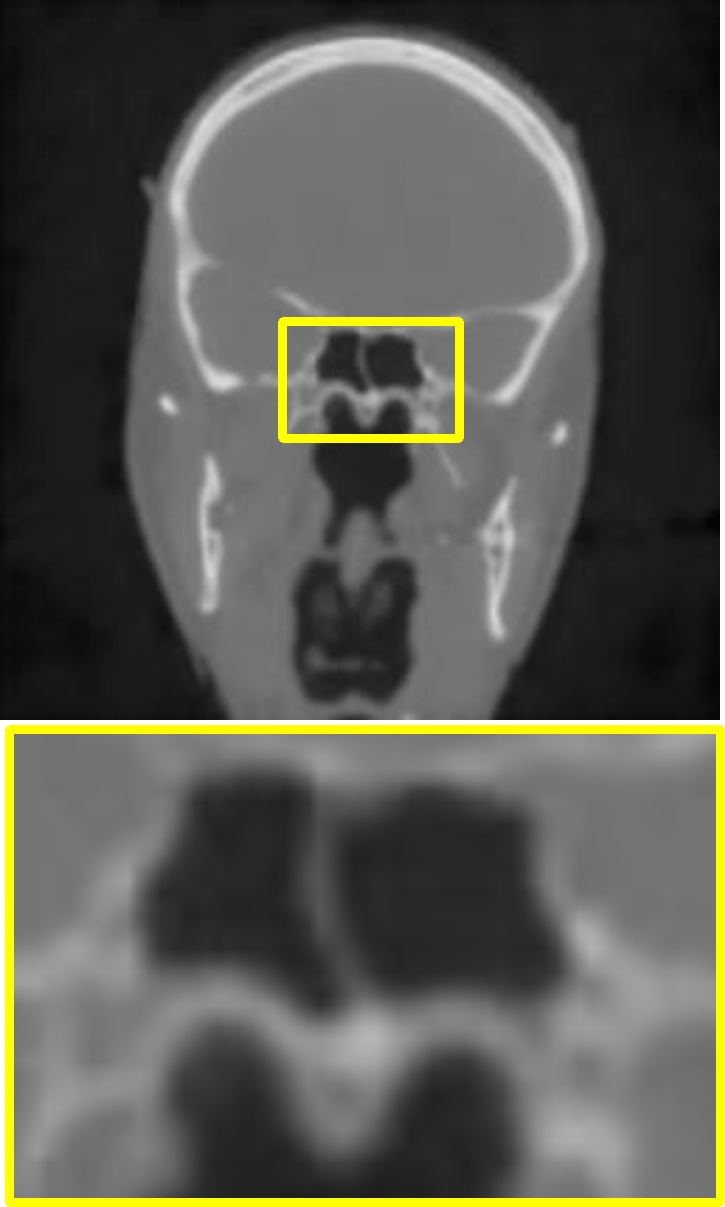}}
        \end{minipage}
        \begin{minipage}[b]{0.12\linewidth}
            \centering
            \centerline{\includegraphics[width=\linewidth]{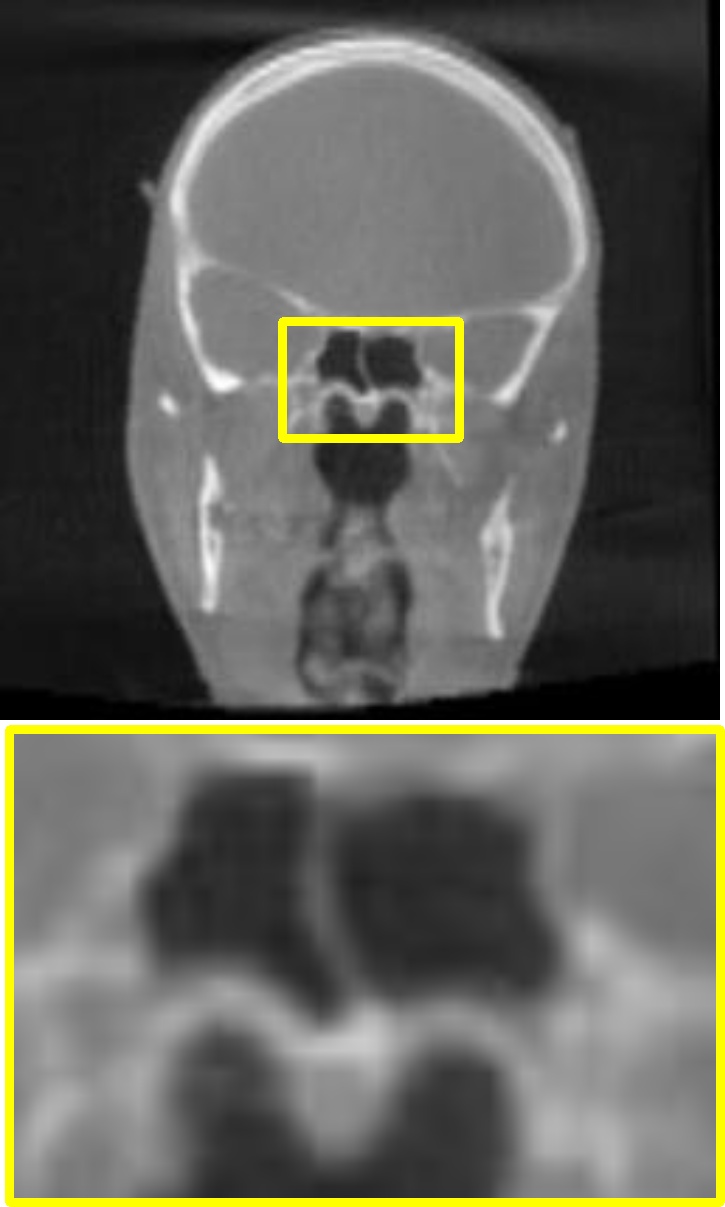}}
        \end{minipage}
    \end{minipage}

    \begin{minipage}[b]{1.0\linewidth}  
        \begin{minipage}[b]{0.12\linewidth}
            \centering
            \centerline{\includegraphics[width=\linewidth]{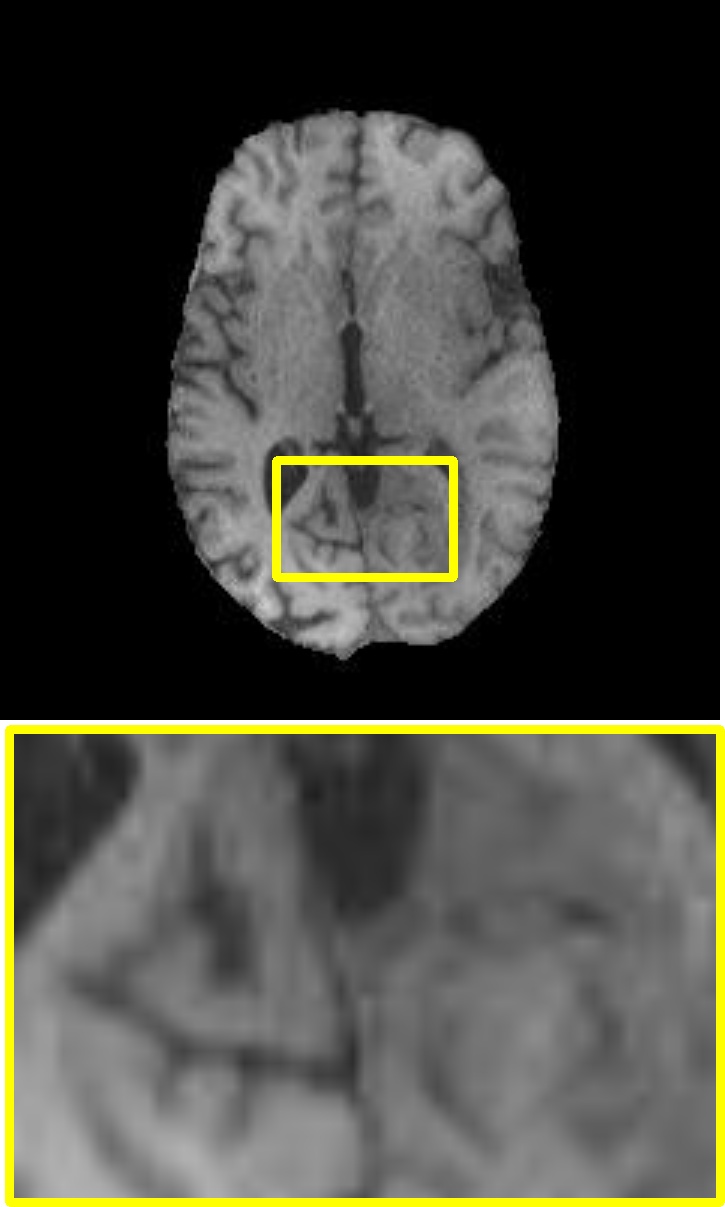}}
        \end{minipage}
        \begin{minipage}[b]{0.12\linewidth}
            \centering
            \centerline{\includegraphics[width=\linewidth]{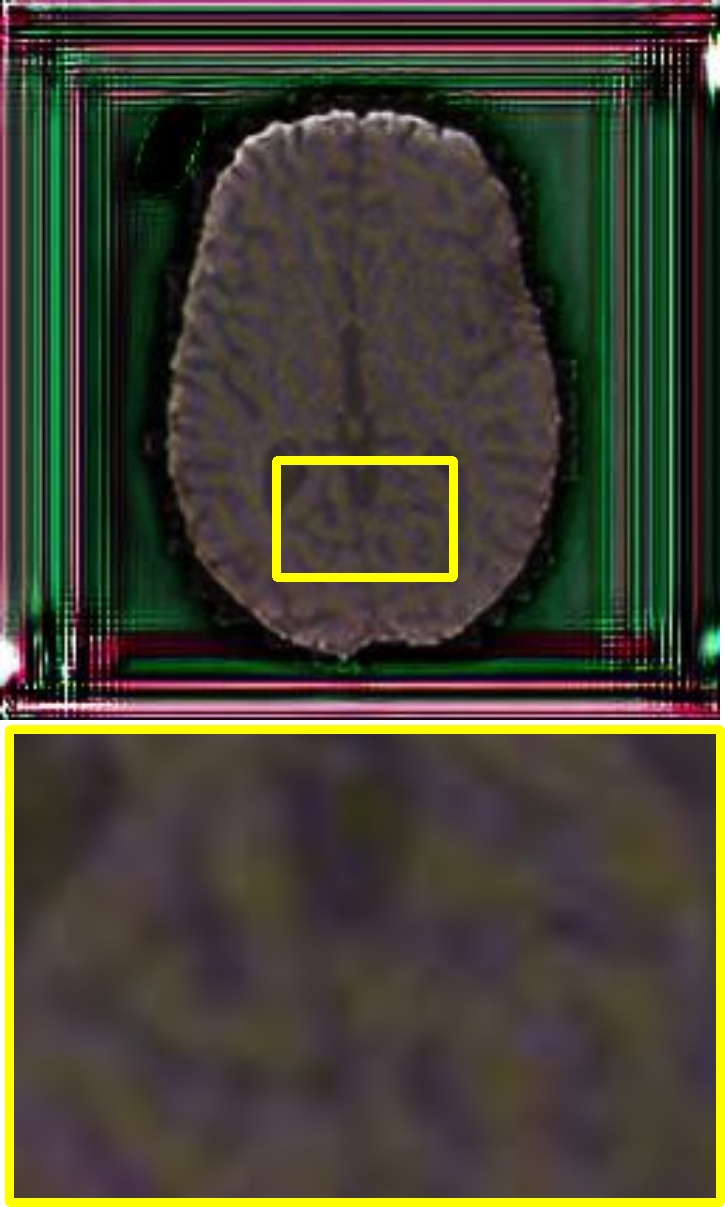}}
        \end{minipage}
        \begin{minipage}[b]{0.12\linewidth}
            \centering
            \centerline{\includegraphics[width=\linewidth]{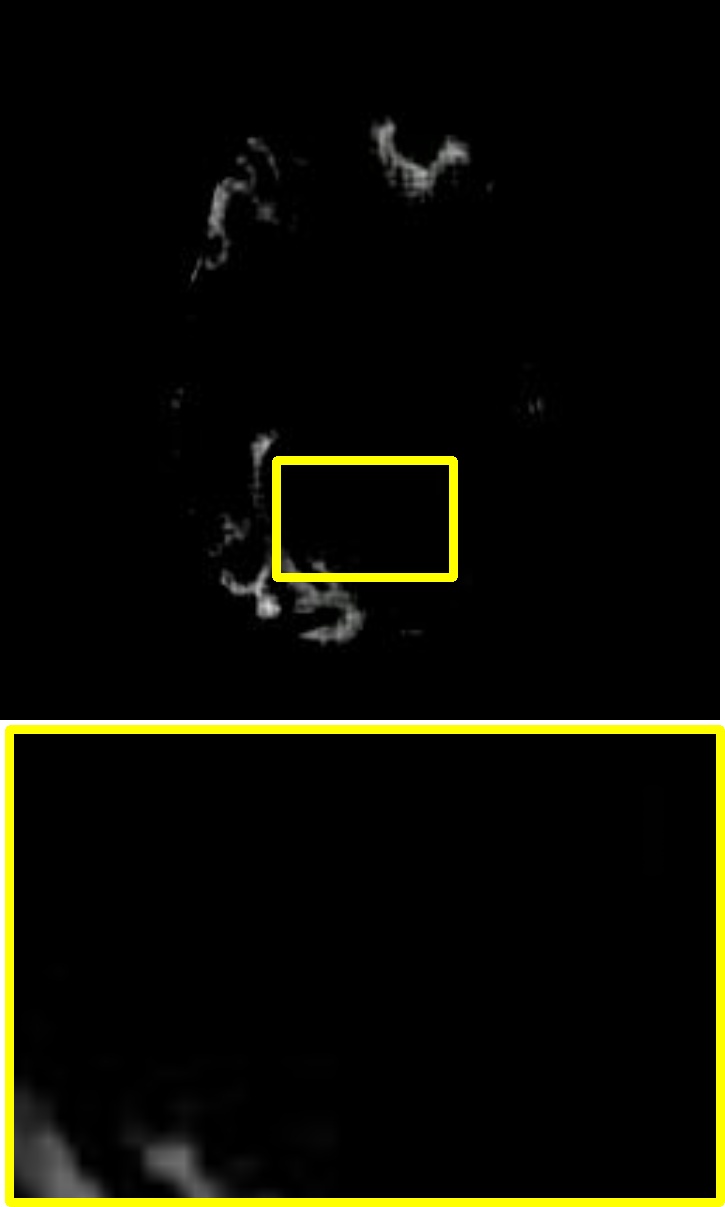}}
        \end{minipage}
        \begin{minipage}[b]{0.12\linewidth}
            \centering
            \centerline{\includegraphics[width=\linewidth]{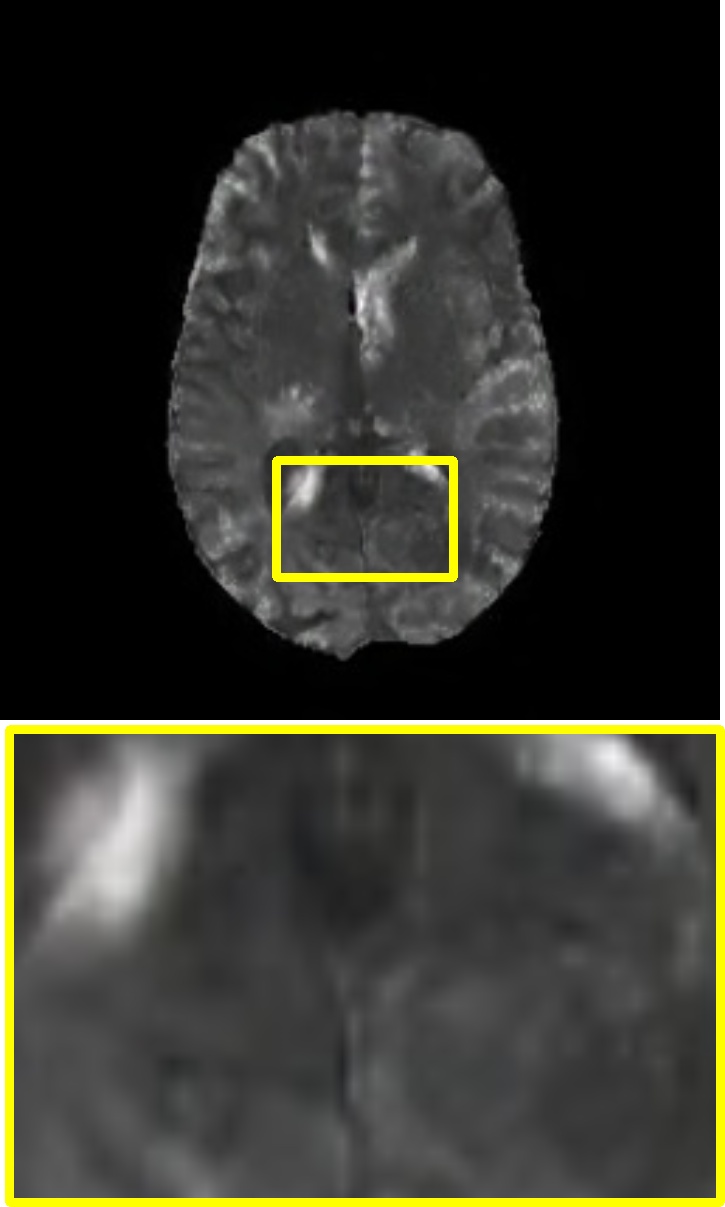}}
        \end{minipage}
        \begin{minipage}[b]{0.12\linewidth}
            \centering
            \centerline{\includegraphics[width=\linewidth]{figs/Bra-T2/mag_lptn.jpg}}
        \end{minipage}
        \begin{minipage}[b]{0.12\linewidth}
            \centering
            \centerline{\includegraphics[width=\linewidth]{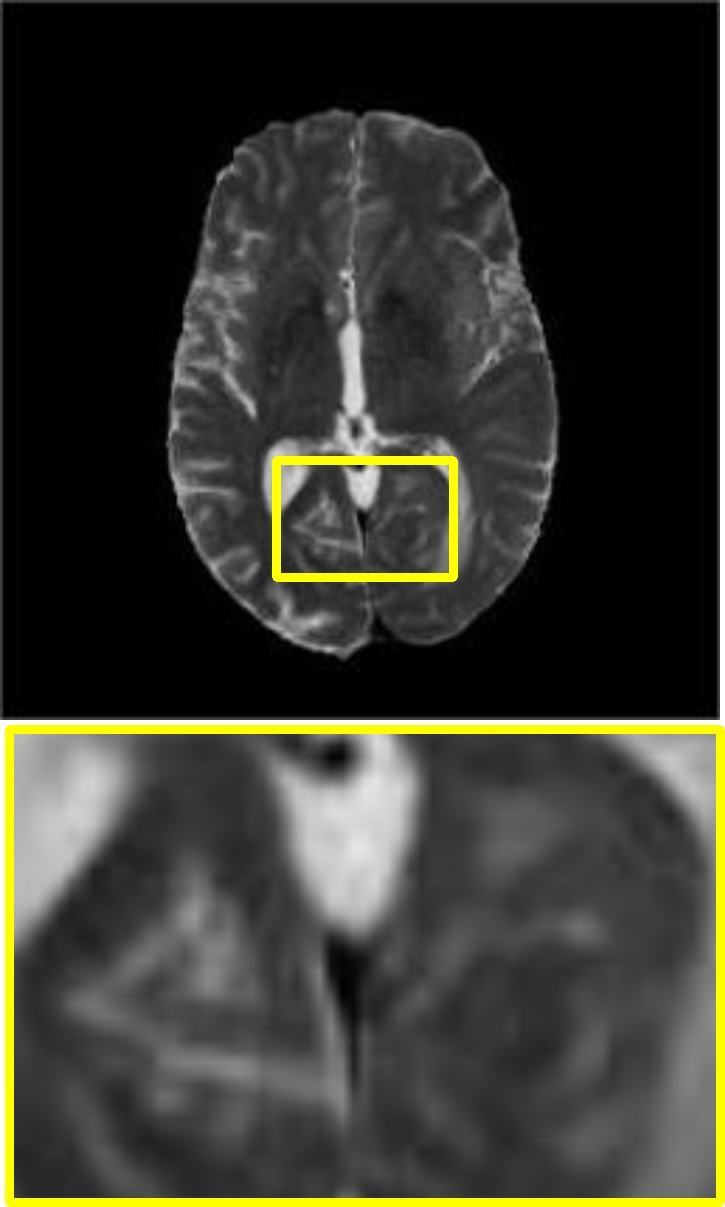}}
        \end{minipage}
        \begin{minipage}[b]{0.12\linewidth}
            \centering
            \centerline{\includegraphics[width=\linewidth]{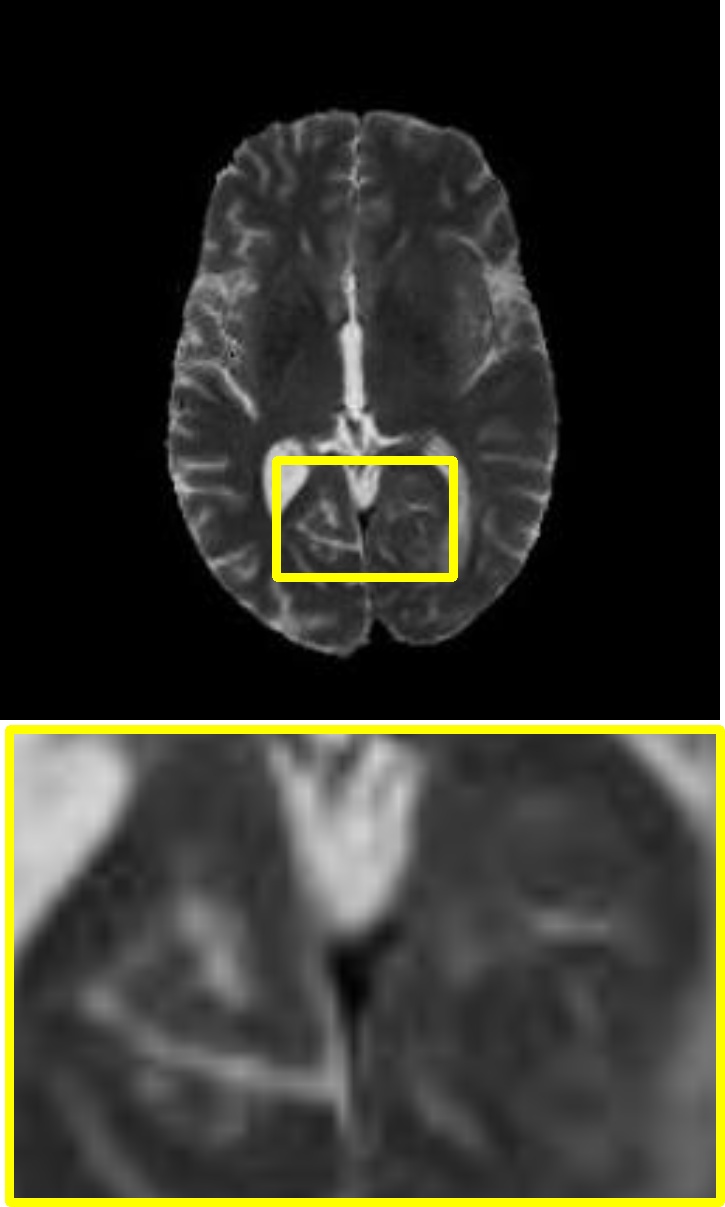}}
        \end{minipage}
        \begin{minipage}[b]{0.12\linewidth}
            \centering
            \centerline{\includegraphics[width=\linewidth]{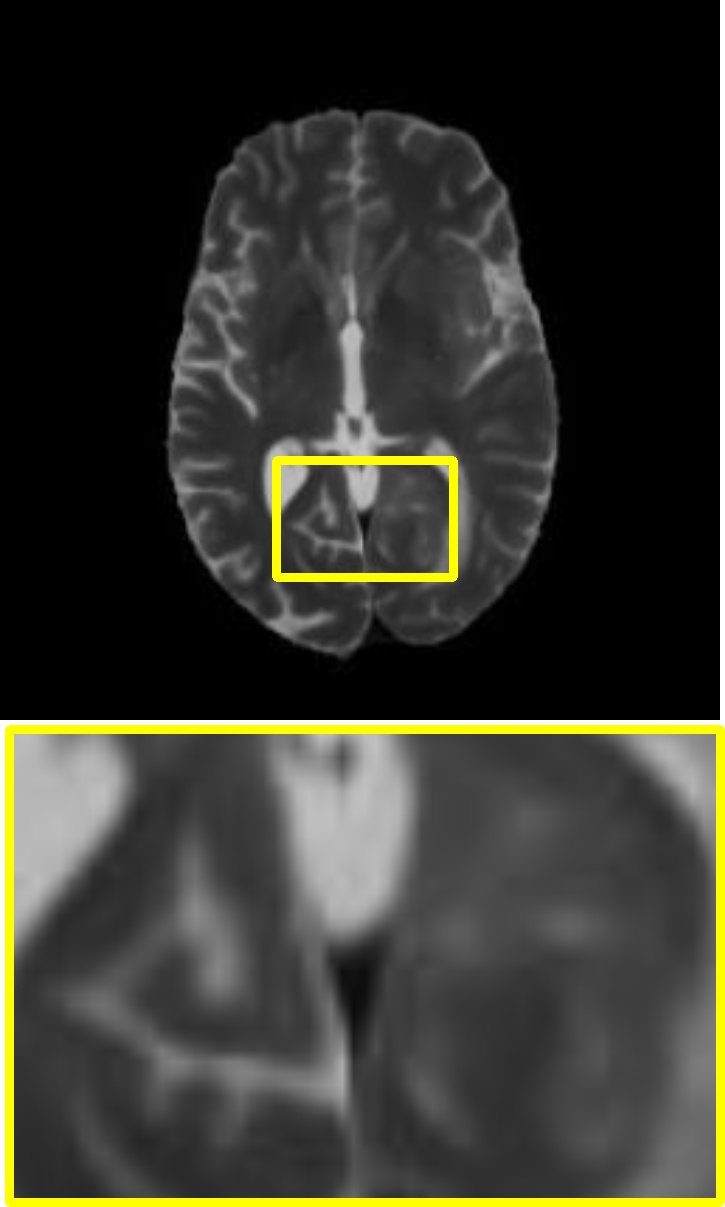}}
        \end{minipage}
    \end{minipage}
    
    \begin{minipage}[b]{1.0\linewidth}
        \begin{minipage}[b]{0.12\linewidth}
            \centering
            \centerline{\includegraphics[width=\linewidth]{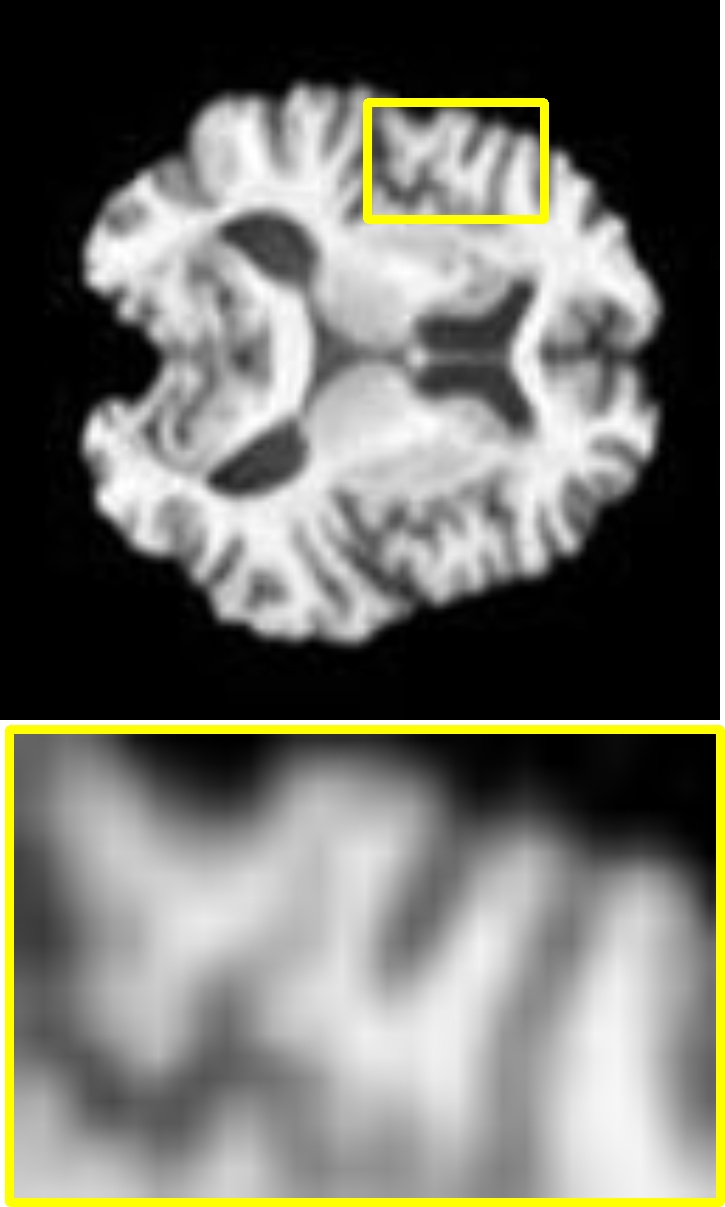}}
            \centerline{(a) Input}\medskip
        \end{minipage}
        \begin{minipage}[b]{0.12\linewidth}
            \centering
            \centerline{\includegraphics[width=\linewidth]{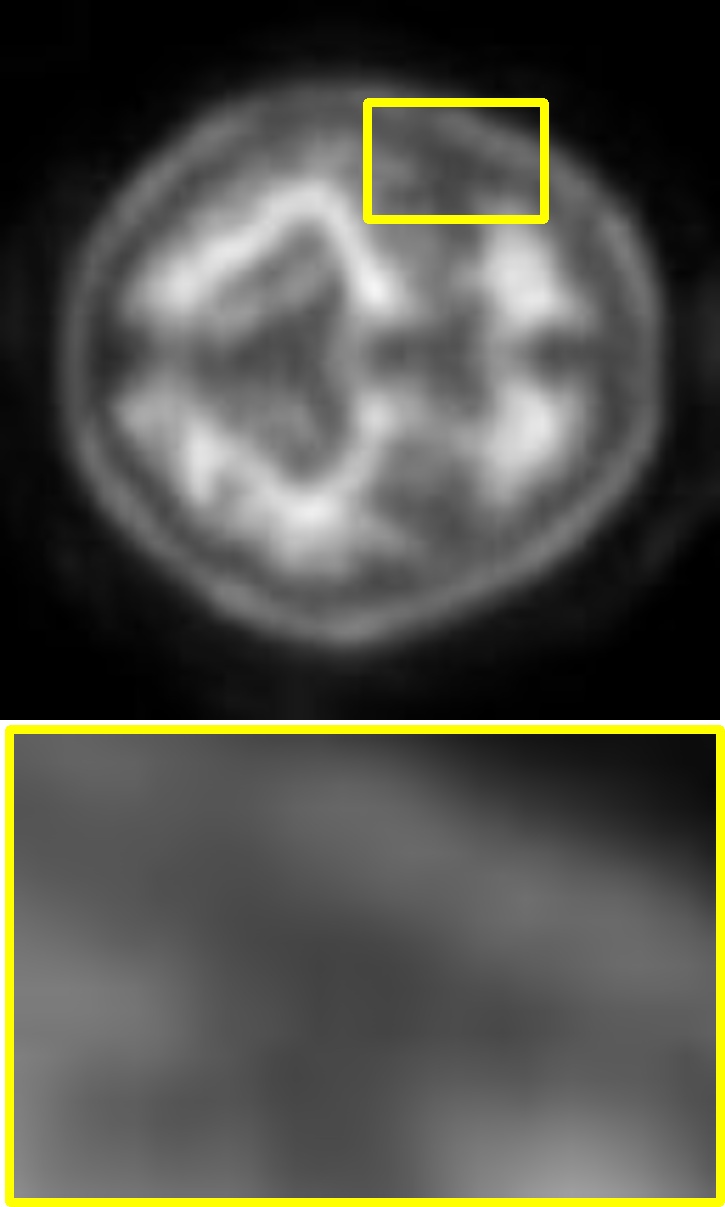}}
            \centerline{(b) CUT}\medskip
        \end{minipage}  
        \begin{minipage}[b]{0.12\linewidth}
            \centering
            \centerline{\includegraphics[width=\linewidth]{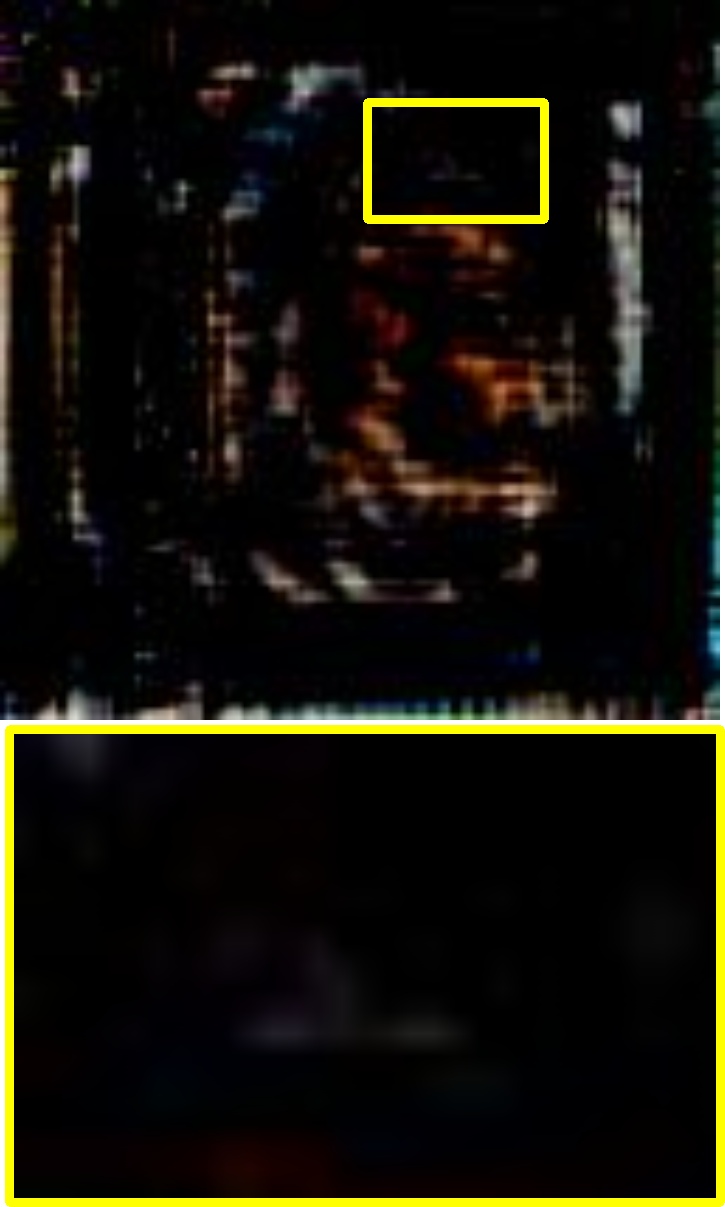}}
            \centerline{(c) CycleGAN}\medskip
        \end{minipage}   
        \begin{minipage}[b]{0.12\linewidth}
            \centering
            \centerline{\includegraphics[width=\linewidth]{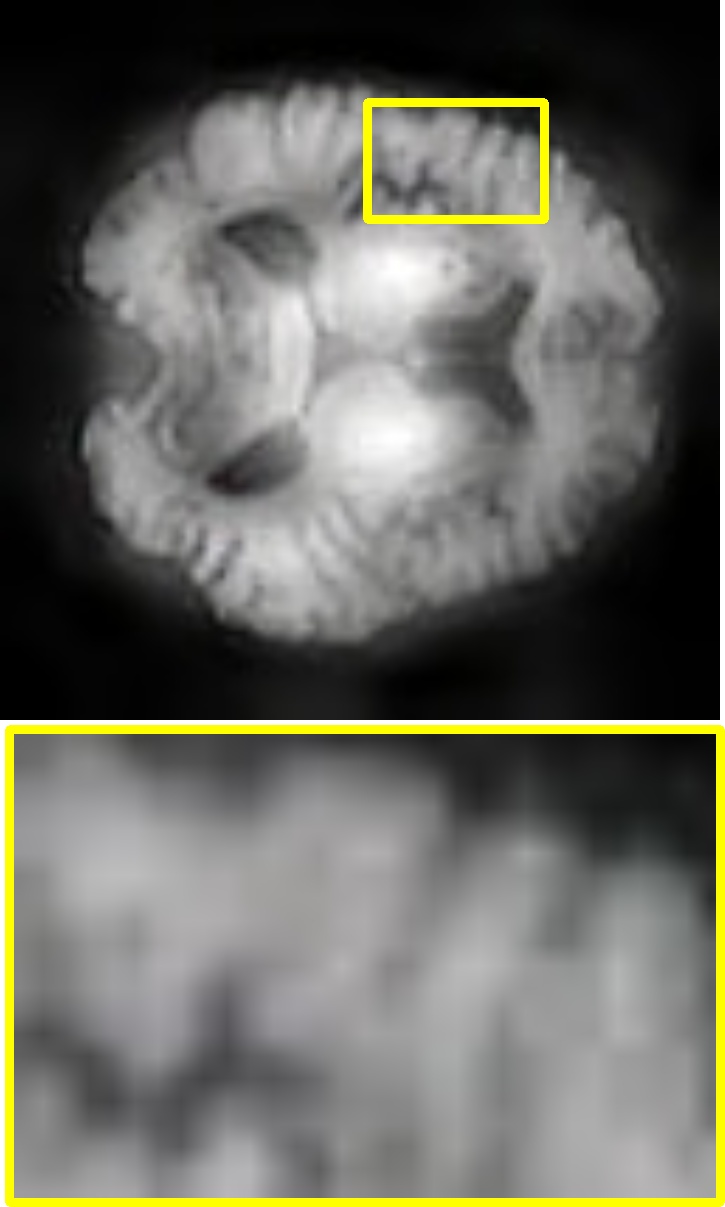}}
            \centerline{(d) LPTN}\medskip
        \end{minipage}
        \begin{minipage}[b]{0.12\linewidth}
            \centering
            \centerline{\includegraphics[width=\linewidth]{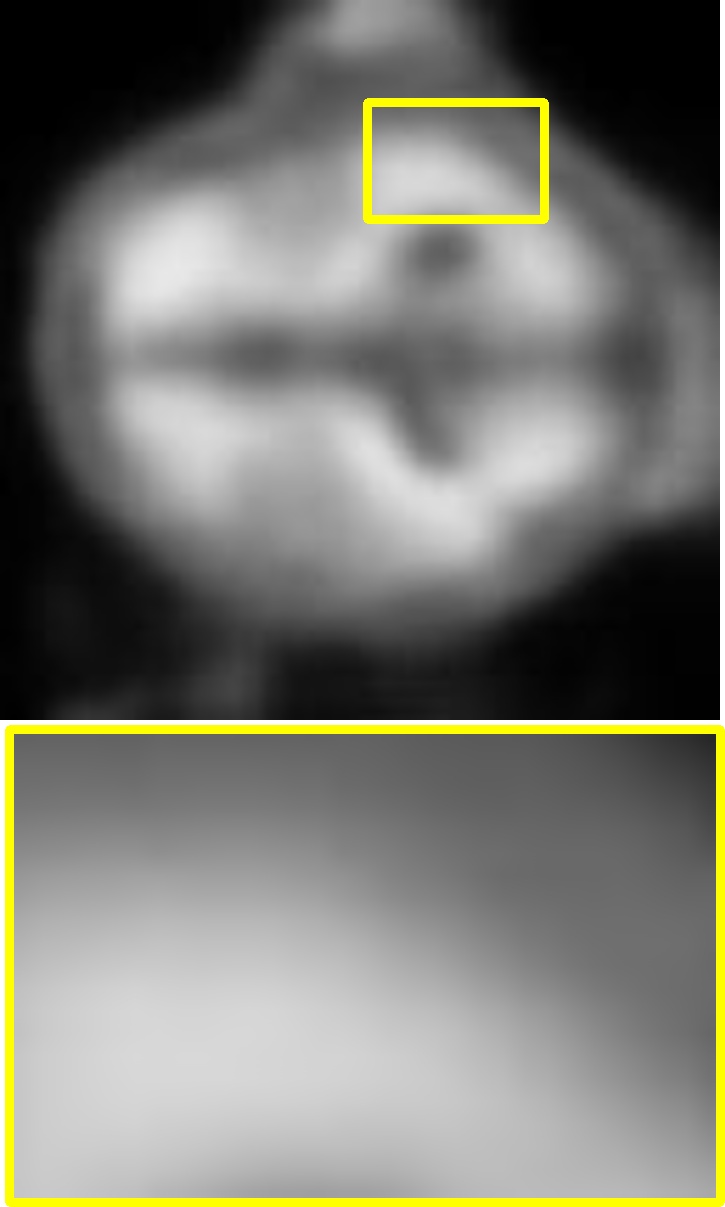}}
            \centerline{(e) pGAN}\medskip
        \end{minipage}
        \begin{minipage}[b]{0.12\linewidth}
            \centering
            \centerline{\includegraphics[width=\linewidth]{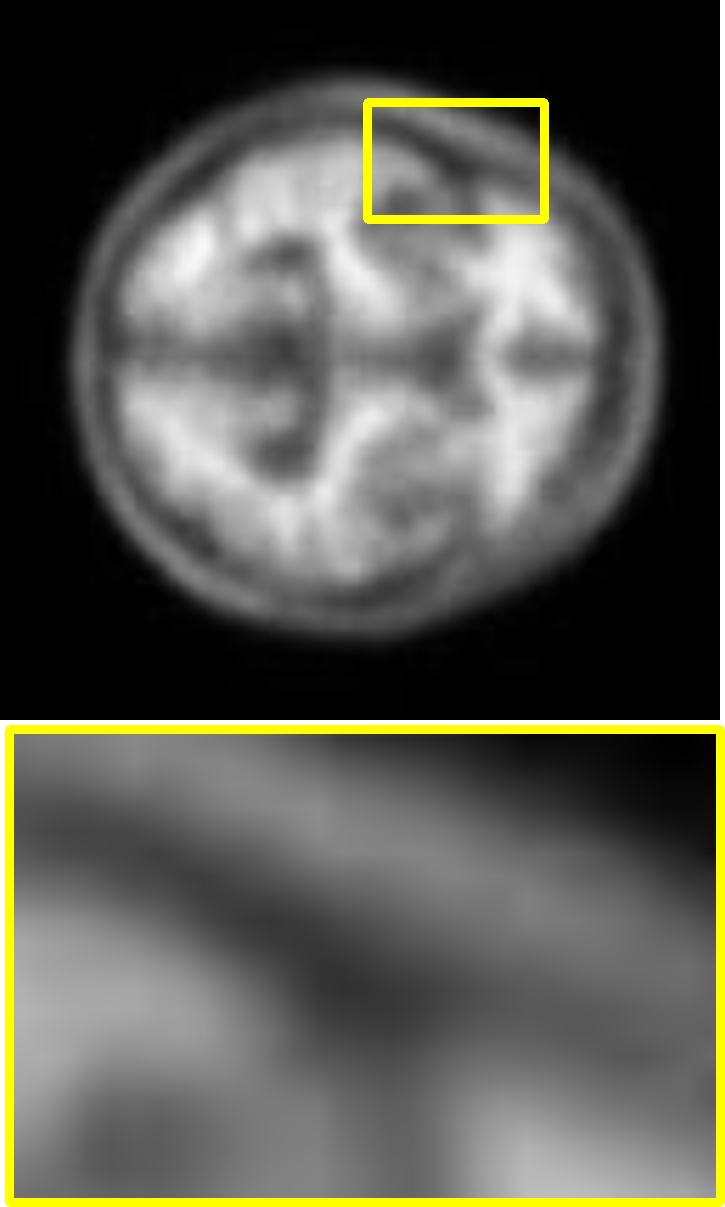}}
             \centerline{(f) ResViT}\medskip
        \end{minipage}
        \begin{minipage}[b]{0.12\linewidth}
            \centering
            \centerline{\includegraphics[width=\linewidth]{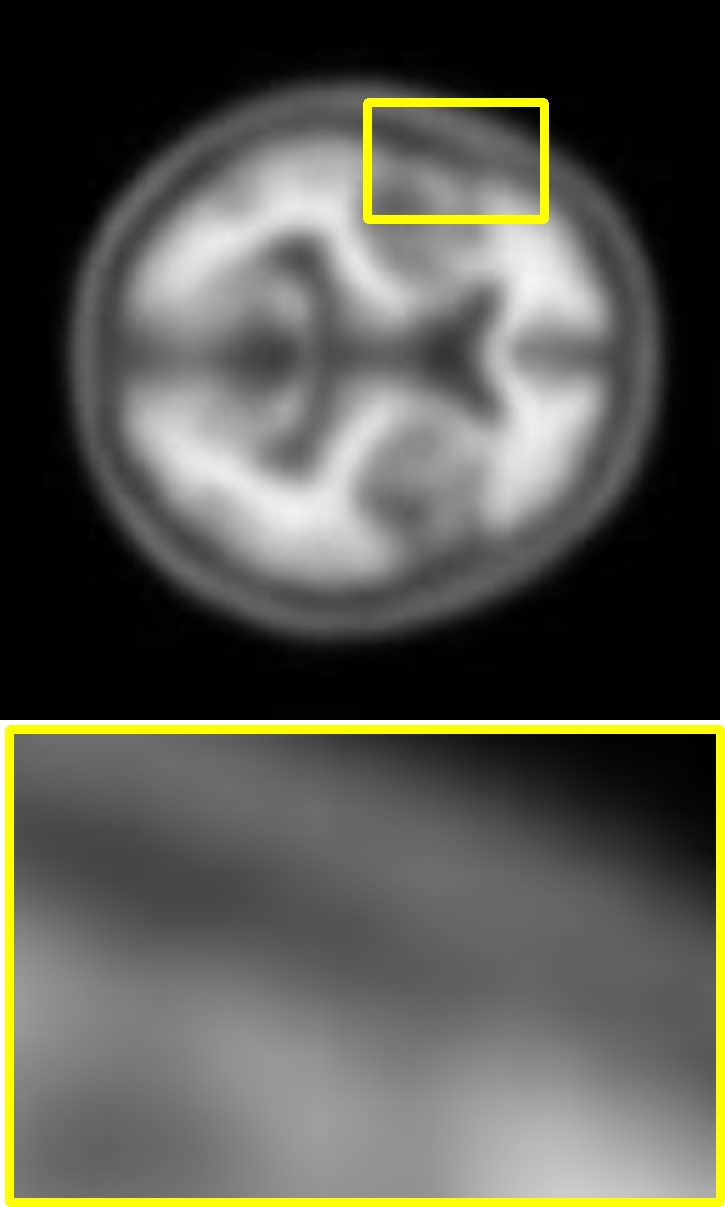}}
            \centerline{(g) Ours}\medskip
        \end{minipage}
        \begin{minipage}[b]{0.12\linewidth}
            \centering
            \centerline{\includegraphics[width=\linewidth]{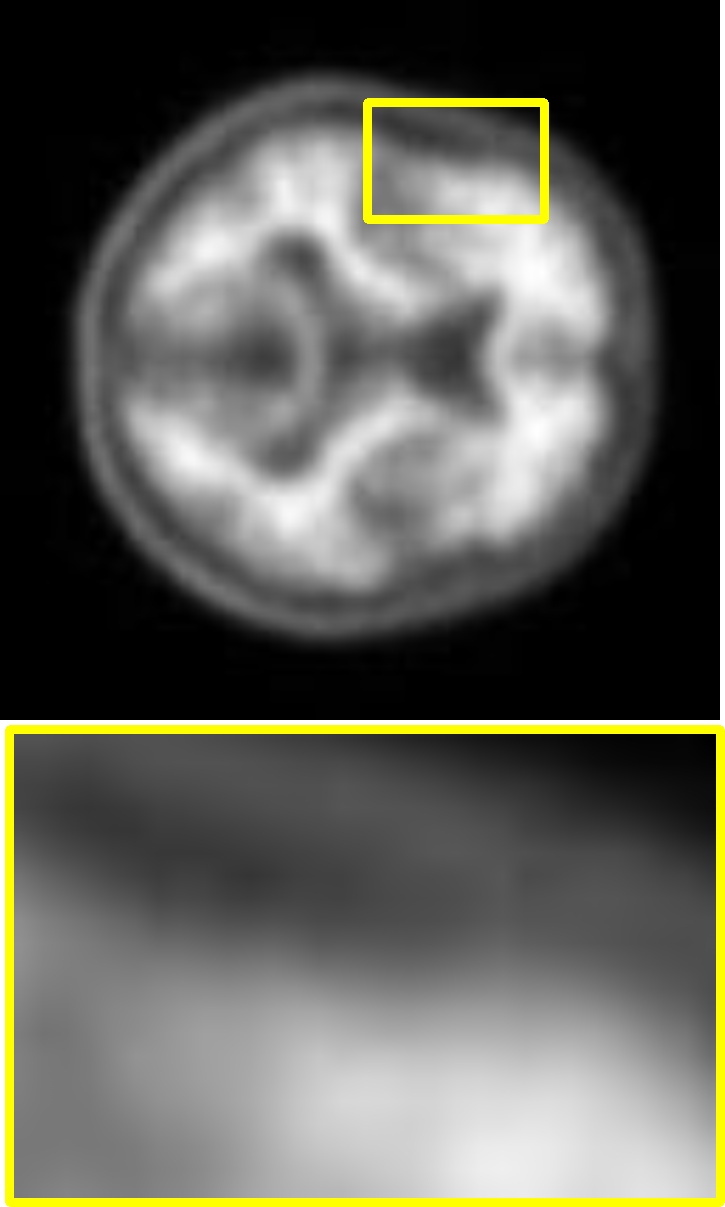}}
            \centerline{(h) Target}\medskip
        \end{minipage}
    \end{minipage}
    
    \caption{
    A visual comparison of different image enhancement methods was conducted on the five distinct datasets. The first row represents the MRI to CT transformation from the SynthRAD dataset, the second row shows the T2 to T1 transformation from the IXI dataset, the third row depicts the MRI to PET transformation from the ADNI dataset, the fourth row displays the CT to CBCT transformation from the SynthRAD dataset, and the last row represents the T1 to T2 transformation from the BraTS dataset. We can clearly observe that CUT (b) and CycleGAN (c) exhibit poor performance in multi-modal image translation. LPTN (d) and pGAN (e) perform relatively better, while ResViT (f) demonstrates the best performance but still falls short in some aspects. Our proposed method (g) successfully reconstructs the target with good fidelity in terms of both details and shape.
    }
    \label{fig:res}
\end{figure*}

The visual results are shown in Figure~\ref{fig:res}.We can observe that CycleGAN (c) performs the worst, as it fails to convert almost all modalities successfully. CUT (b) and LPTN (d) perform relatively better, as they partially succeed in modality conversion, although there are significant differences in details and shape compared to the target (h). pGAN (e) demonstrates relatively successful transformation in the first two modalities but exhibits significant differences in details compared to the target. However, it performs poorly in the last modality. ResViT (f) performs well across all modalities, but there are still certain gaps in detail compared to the target. For example, in the first row there are shape disparities, in the second row there are issues with noise in darker regions, and in the last row the edge is different from the target. Finally, our proposed method (g) performs well across all modalities and exhibits the closest resemblance to the target in terms of details.


\subsection{Ablation Studies}
In this section, we conduct the following ablation experiments on all datasets, the results can be seen in Table~\ref{tabel:ab}:
\begin{enumerate}
    \item \textbf{Ours w/o PEB:} Remove the Prompt Extraction Block.
    \item \textbf{Ours w/o PFB:} Remove the Prompt Fusion Block.
    \item \textbf{Ours w/o Transformer:} Remove the Transformer block.
    \item \textbf{Ours Full:} Our full MedPrompt architecture.
\end{enumerate}

As shown in Table~\ref{tabel:ab}, it is evident that the performance of the network is significantly affected when PEB and PFB are removed. Specifically, the PSNR for each modality decreases by approximately 3 dB, the SSIM decreases by around 0.7 dB, and the MAE increases by about 2 dB. These results indicate that the removal of PEB and PFB has a substantial impact on the performance of the network, which demonstrates that the PEB and PFB play a crucial role in terms of multi-task learning and cross-modal transferring. When the Transformer block is removed, we also observe a certain degree of performance degradation. Specifically, the PSNR for each modality decreases by approximately 2 dB, the SSIM decreases by approximately 0.03 dB, and the MAE increases by approximately 0.5 dB. This phenomenon demonstrates that our simple encoder-decoder Transformer architecture can also make a significant contribution to the network's performance. Additionally, PEB and PFB exhibit greater efficacy when integrated with this simplified Transformer architecture.

\section{Conclusion}


In this paper, we propose MedPrompt, a straightforward yet effective multi-task medical image translation framework. By leveraging the large receptive field of the Transformer and the effective cross-modal feature extraction of prompting, MedPrompt achieves state-of-the-art performance across various pairs of modalities, demonstrating excellent generalization capability. Furthermore, we propose two key components: the Prompt Extraction Block (PEB) and the Prompt Fusion Block (PFB), which selectively extract and aggregate prompt information from different modalities. The PEB generates modality-specific prompt weights, while the PFB dynamically fuses the extracted prompts based on their relevance to the target modality. These components make substantial contributions to multi-task learning. We conduct extensive experiments and demonstrate our method is superior in terms of multi-task performance and convenience which only requires a single training process. Although our proposed framework demonstrates good generalization capability, there is still room for further improvements across different domains. As a next step, we aim to explore and propose more effective prompt methods. 

\bibliography{aaai24}

\end{document}